\documentclass[letterpaper, 10 pt, journal, twoside]{IEEEtran}

\usepackage{decar-common}
\usepackage{decar-dynamics}
\usepackage{decar-lie}
\usepackage{decar-addons}
\usepackage[T1]{fontenc} 
\usepackage{anyfontsize} 

\usepackage[dvipsnames]{xcolor}

\newsavebox{\subfigbox}

\makeatletter
\AtBeginDocument{
  \check@mathfonts
}
\makeatother

\begin{document}



\fontdimen16\textfont2=\fontdimen17\textfont2
\fontdimen13\textfont2=5pt

\title{DIVO: Continuous-time DVL-Inertial-Visual Odometry for Unmanned Underwater Vehicles}

\author{
Kyungmin~Jung, 
Angad~Bajwa, 
Junha~Yoo, 
Arturo~Del~Castillo~Bernal, 
and~James~Richard~Forbes

\thanks{This work was supported in part by Voyis Imaging Inc.\ through the Natural Sciences and Engineering Research Council of Canada (NSERC) Alliance program, the NSERC Discovery Grant program, and in part by the McGill Engineering Doctoral Award (MEDA) Program at McGill University.}%

\thanks{The authors are with the Department of Mechanical Engineering, McGill University, Montreal, QC, Canada, H3A 0C3. \\
{\tt\footnotesize \{kyungmin.jung, angad.bajwa, junha.yoo, arturo.delcastillobernal\}@mail.mcgill.ca, \\ james.richard.forbes@mcgill.ca}}%

}


\maketitle

\begin{abstract}
    This paper presents a novel acoustic-visual-inertial odometry solution leveraging a continuous-time trajectory estimation framework for unmanned underwater vehicles.
    Underwater environments present unique challenges for visual localization and mapping, such as light attenuation, illumination variance, and the presence of particulate matter.
    This motivates the use of additional sensing modalities and a visual tracking pipeline that is robust to diverse subsea conditions.
    The proposed system is the first continuous-time trajectory estimation framework based on Gaussian processes to fuse asynchronous measurements from a Doppler velocity log, a stereo camera, and an inertial measurement unit. Additionally, a novel visual frontend is proposed, incorporating learning-based feature extraction and matching that is robust to the specific challenges that subsea environments present. 
    The proposed framework enables seamless integration of additional sensor modalities in continuous-time and is adaptable to different environments without reconfiguration.
    The proposed system is extensively tested on real-world underwater inspection datasets, where it outperforms state-of-the-art visual-inertial and acoustic-visual-inertial SLAM algorithms in accuracy, robustness, and trajectory coverage.
    Notably, the proposed system outperforms the state-of-the-art despite only forming short-term visual data associations.
    
\end{abstract}


\section{Introduction}
\label{sec:intro}

\begin{figure}[ht!]
    \centering
    \includegraphics[width=\linewidth]{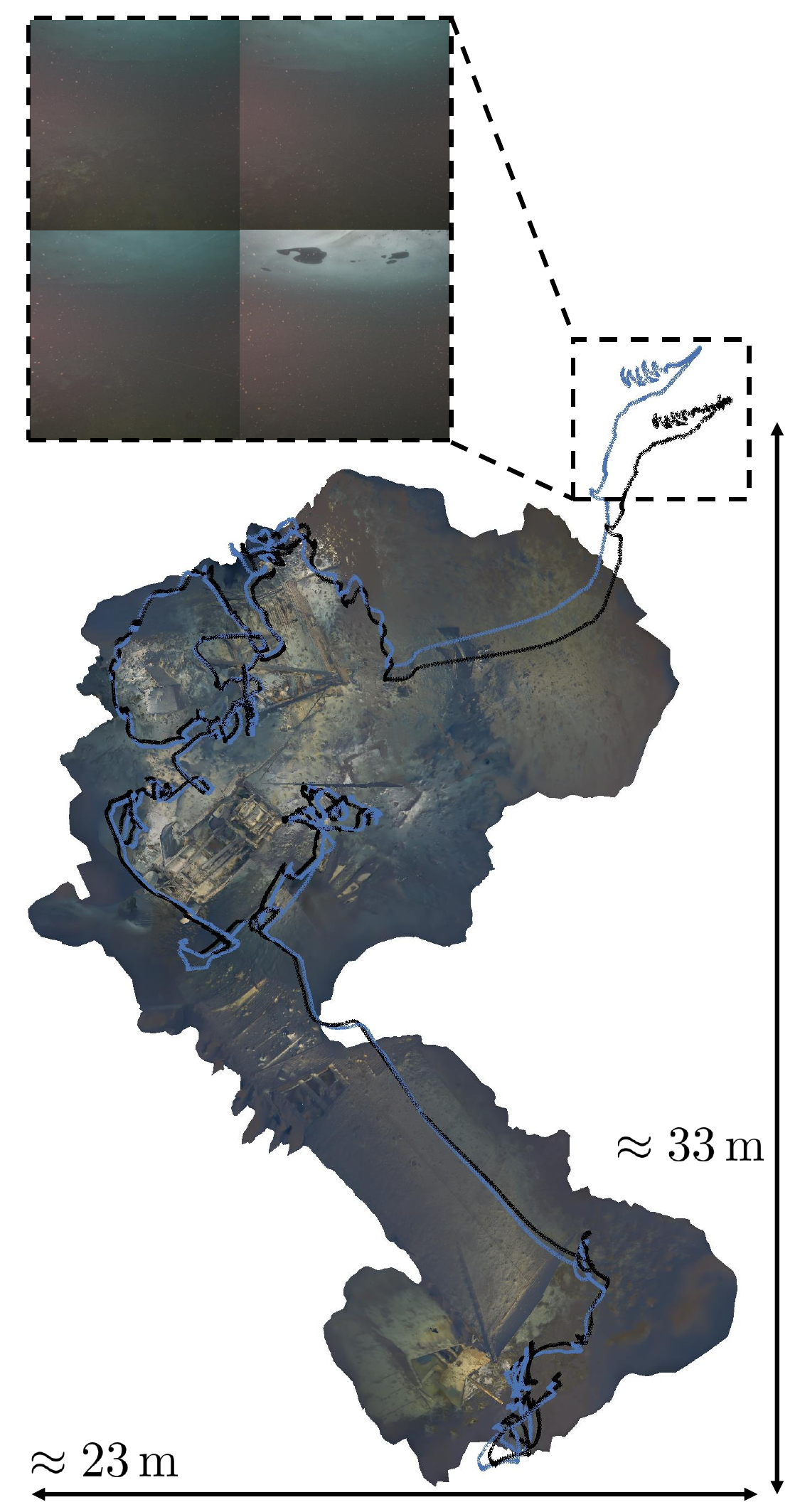}
    \caption{
        The ROV's trajectory estimated by the proposed DIVO method (blue) is overlaid on top of the ground truth trajectory (black) and the mesh generated using Agisoft Metashape.
        The proposed system can robustly register images in visually challenging scenes such as low visibility and high presence of dynamic particles. As an example, a sample of images at the beginning of the trajectory are shown. 
    }
    \label{fig:divo_map}
\end{figure}
\IEEEPARstart{U}{nmanned} underwater vehicles (UUVs) have been widely used in underwater inspection, mapping, and exploration~\cite{Paull2014AUV}.
The success of these missions relies heavily on accurate and robust localization and mapping capabilities.
Considering that most UUVs are equipped with cameras, visual simultaneous localization and mapping (SLAM) techniques have been extensively studied for underwater applications~\cite{Negahdaripour2006ROV, Eustice2008Visually, Pizarro2009Large}.

Despite great success above water~\cite{MurArtal2015ORBSLAM, Qin2018VINS-Mono, Engel2018DSO}, visual-only SLAM faces unique challenges in underwater environments.
Poor visibility caused by light attenuation limits the range of visual feature detection, dynamic lighting conditions due to caustics and turbidity cause tracking failure, and the presence of particulate matter such as suspended sediments and organic detritus, known as ``marine snow", can occlude and distract visual features.
Their performance is further compromised in feature-scarce environments such as open water and seabeds~\cite{Eustice2008Visually}.

To overcome these challenges, fusing visual data with inertial measurement unit (IMU) data has been proposed and shown to improve the robustness of underwater visual SLAM~\cite{Hidalgo2020ORBSLAM2, Singh2025Aeriel}.
Visual-inertial SLAM leverages the ego-motion constraints provided by the IMU to aid pose estimation, but only functions under visual failure for a short period of time due to the unbounded drift of the inertial navigation solution. 
Additionally, when tested on subsea data, many of the above water state-of-the-art algorithms show significant non-determinism~\cite{Joshi2019Experimental}.

Historically, the use of acoustic sensors, such as a Doppler velocity log (DVL)~\cite{Rahman2019SVIn2, Song2024TURTLMap}, has been a popular complementary sensor used to bound the drift of the inertial navigation solution over time.
The DVL provides velocity measurements relative to the seabed by measuring the Doppler shift of acoustic beams reflected off the seabed.
Fusing DVL and IMU data has seen success in improving the localization robustness~\cite{Potokar2021Invariant, Song2023Uncertainty, Song2024TURTLMap}.
However, fusion of the DVL and IMU without an additional exteroceptive sensor still leads to a drifting navigation solution~\cite{Maurelli2022Auv}.

Recently, there has been a growing interest in fusing DVL, camera, and IMU measurements to leverage the complementary strengths of each sensor modality~\cite{Rahman2019SVIn2, Thomas2023Tightly, Xu2025AQUA-SLAM}.
However, fusing multiple sensors with different sampling rates poses additional challenges.
To address these problems, Zhao \etal~\cite{Zhao2023Tightly} presented a multi-sensor fusion framework using a multi-state constraint Kalman filter.
Later, Xu \etal~\cite{Xu2025AQUA-SLAM} proposed a SLAM system that couples DVL, camera, and IMU measurements in an optimization-based framework by employing the joint preintegration of IMU and linear velocity measurements~\cite{Vial2025Lie}.
Although this approach effectively handles the fusion of IMU and linear velocity measurements, modular fusion of additional asynchronous sensor modalities remains an open problem, necessitating the use of a continuous-time backend~\cite{Talbot2025Continuous}.

The problems of poor subsea visual tracking and asynchronous sensor measurements are addressed in this paper by combining a novel learned visual frontend and a continuous-time backend that supports asynchronous multi-sensor fusion. The main contributions of this paper are
\vspace{-0.2cm}
\begin{itemize}
    \item a novel acoustic-visual-inertial odometry system that fuses asynchronous DVL, camera, and IMU measurements in a continuous-time trajectory estimation framework using Gaussian processes~\cite{Barfoot2014BatchCT, Anderson2015STEAM},
    \item a novel learned visual frontend using SuperPoint~\cite{detoneSuperPointSelfSupervisedInterest2018} features matched with LightGlue~\cite{lindenbergerLightGlueLocalFeature2023}, and
    \item extensive evaluation on real-world underwater inspection datasets with metric photogrammetric ground truth.
\end{itemize}
The proposed system, \textbf{D}VL-\textbf{I}nertial-\textbf{V}isual-\textbf{O}dometry (DIVO), is the first continuous-time optimization-based odometry system designed for UUVs. Additionally, DIVO is the first underwater odometry system to incorporate a learned frontend for feature matching. 
The proposed method has been tested in both simulation and extensive real-world experiments. 
It outperforms the state-of-the-art of both visual-inertial and acoustic-visual-inertial domains in estimation accuracy, trajectory coverage, and robustness, despite using only short-term visual data associations.

The remainder of this paper is organized as follows.
Section~\ref{sec:relatedwork} presents a comprehensive review of the literature on DVL-aided navigation, visual and subsea SLAM, and continuous-time estimation, followed by the preliminaries of continuous-time trajectory estimation in Section~\ref{sec:continuousestimation}.
Section~\ref{sec:divo} presents the derivation of the proposed approach, followed by its experimental evaluation in Section~\ref{sec:results}.
Finally, the paper is drawn to a close in Section~\ref{sec:conclusion}.


\section{Related Work}
\label{sec:relatedwork}
In this section, relevant literature as it pertains to DVL-aided navigation, underwater SLAM, and continuous-time estimation is reviewed.

\subsection{DVL-aided Navigation}
\label{subsec:relatedwork-dvl}
Historically, the use of Doppler velocity logs has been restricted to larger UUVs that have the payload capacity and data bandwidth to carry the transducer arrays~\cite{Snyder2010DVL}.
Recently, smaller DVLs have become increasingly popular as navigation tools for smaller autonomous underwater vehicles (AUVs) and remotely operated vehicles (ROVs)~\cite{Pan2025Novel}.

Both loosely-coupled and tightly-coupled methodologies have been proposed to fuse DVL and IMU data for the purposes of underwater navigation.
In loosely-coupled approaches, the DVL velocity is deduced from the transducer measurements and fused with the IMU using, for example, a Kalman filter~\cite{Kinsey2006Survey, Luo2022SINS}.
However, the main drawback of a loosely-coupled approach is that the DVL must have at least three transducers in bottom lock to infer velocity, which may not always be possible in deep waters or over unstructured seabeds.
To overcome this limitation, Tal \etal~\cite{Tal2017Inertial} proposed a tightly-coupled approach, where raw DVL beam measurements are directly compared with the estimated velocities computed from an IMU.

Recently, optimization-based methods have been employed to fuse DVL data with other sensors~\cite{Thomas2023Tightly, Song2024TURTLMap}.
The preintegration of linear velocity measurements obtained by a DVL is combined with angular velocity measurements from an IMU as binary factors encoding relative position in a decoupled mapping for infrastructure inspection~\cite{Thomas2023Tightly}.
Additionally, Vial \etal~\cite{Vial2025Lie} proposed the preintegration of IMU and linear velocity measurements that sets a new motion constraint in ${SE_{N}(3)}$ in optimization-based SLAM problems.
Song \etal~\cite{Song2024TURTLMap} used this preintegration pipeline and proposed a solution that focuses on real-time localization based on DVL, IMU, and a barometer with a decoupled visual mapping for textureless environments.

\subsection{Visual SLAM}
\label{subsec:relatedwork-vslam}
Visual SLAM is one of the most mature substrata of the overall SLAM literature.
Camera data is commonly fused with some interoceptive sensor modality, such as an IMU. Estimating motion from RGB images and generating useful map representations for downstream tasks are objectives that continue to see focus in the literature.
The methods here can be roughly demarcated into loosely versus tightly coupled, and recursively filtering a subset of states, e.g.~\cite{Geneva2020OpenVINS}, versus optimizing a window of states, e.g.~\cite{Leutenegger2015Keyframe}.
These methods can further be demarcated into direct versus indirect and sparse versus dense.
Of all combinations including direct-sparse~\cite{Engel2018DSO} and indirect-dense~\cite{ranftlDenseMonocularDepth2016}, popular approaches are keypoint-based indirect-sparse methods VINS-Fusion~\cite{Qin2018VINS-Mono} and ORB-SLAM3~\cite{camposORBSLAM3AccurateOpenSource2021}.
Recently, OKVIS2-X~\cite{bocheOKVIS2XOpenKeyframebased2025} has introduced dense volumetric mapping into the estimation pipeline through map-to-map and frame-to-map alignment factors.

Though traditional geometric feature extraction and matching methods such as the KLT sparse optical flow algorithm~\cite{Shi1994Good, Lucas1981KLT} remain ubiquitous, recent inroads have been made into incorporating deep-learning-based feature methods into visual SLAM pipelines.
The introduction of SuperPoint~\cite{detoneSuperPointSelfSupervisedInterest2018} which uses self-supervised learning with CNN-based features, and DISK~\cite{diskFeature2020}, which uses reinforcement learning to perform end-to-end matching, marked the introduction of learned point features that are suitable for visual SLAM. XFeat~\cite{potjeXFeatAcceleratedFeatures2024} was later introduced as a CNN-based extraction-matching pipeline. However, XFeat features are described by 64-dimensional float vector, as compared to SuperPoint's 256-dimensional float vector.
Learning-based feature matchers such as SuperGlue~\cite{sarlinSuperGlueLearningFeature2020}, which couples positional patterns and photometric context to achieve state-of-the-art matching performance, have also grown in popularity.
However, SuperGlue faces challenges in low-texture settings and shows limited generalizability~\cite{jiangOmniGlueGeneralizableFeature2024}.
LightGlue~\cite{lindenbergerLightGlueLocalFeature2023} introduces early stopping and dynamic network width to the SuperGlue architecture, increasing its flexibility in out-of-domain testing and making it suitable for lightweight deployment.
OmniGlue~\cite{jiangOmniGlueGeneralizableFeature2024} generalizes feature matching by decoupling positional patterns from point feature matching, but is not feasible for real-time visual SLAM applications. In particular, SuperPoint's high dimensionality makes it more adaptable in unstructured environments and to illumination variance, and LightGlue's computational efficiency makes it well-suited to alleviate the processing bottlenecks in real-time SLAM. As such, SuperPoint and LightGlue are selected as the extraction-matcher pair in this work.

Many approaches have explored incorporating learned feature extraction and matching methods into a visual-inertial SLAM pipeline, or directly regressing the camera pose through end-to-end learning.
SuperVINS~\cite{luoSuperVINSRealTimeVisualInertial2025} incorporates SuperPoint-LightGlue as an extension of VINS-Fusion, but they only show comparisons on the EuRoC Dataset~\cite{Burri2016EuRoC}.
SupAtten-SLAM~\cite{liSupAttenSLAMSuperPointVariant2025} integrates an enhanced SuperPoint variant into ORB-SLAM2~\cite{MurArtal2017ORBSLAM2}, but they do not employ a learning-based feature matcher and again evaluate only on indoor environments. Hence, the generalizability of learning-based extractors and matchers to challenging underwater environments still remains unexplored. It is posited that learning-based feature tracking pipelines would generalize well to underwater environments due to their leveraging of global image context and robustness to outliers, a claim that is tested in this work. End-to-end learning-based approaches such as Trianflow~\cite{zhaoBetterGeneralizationJoint2020} and DF-VO~\cite{zhanVisualOdometryRevisited2020} are tested on underwater data in~\cite{alvarez-tunonMonocularVisualSimultaneous2024} but are outperformed by classical methods due to a lack of uncertainty characterization and representative training data. Additionally, visual end-to-end approaches cannot currently be tightly coupled with other sensing modalities, limiting their performance.

\subsection{Underwater Visual SLAM}
\label{subsec:relatedwork-uvslam}
Early visual SLAM approaches were based on fusing navigation data with camera-based 5 DOF relative pose measurements in a loosely-coupled manner~\cite{Negahdaripour2006ROV, Eustice2008Visually}.
The fundamental goal was to concurrently estimate vehicle poses given low-overlap imagery.
Later, subsea-focused SLAM algorithms were extended to incorporate other sensor modalities like DVL in a sparse point-cloud based on a piecewise-planar model~\cite{Ozog2013Real} and in a pose graph framework with DVL modeled as odometry constraints~\cite{Kim2013Real}.
Since then, the enhancement of accuracy and robustness of underwater visual SLAM has been an active research area.

With a plethora of advancements in visual-inertial SLAM techniques, researchers have explored the use of these visual SLAM algorithms for underwater applications~\cite{Joshi2019Experimental}.
Joshi \etal~\cite{Joshi2019Experimental} stated that OKVIS~\cite{Leutenegger2015Keyframe}, SVO~\cite{Forster2017SVO}, ROVIO~\cite{Bloesch2017ROVIO}, and VINS-Mono~\cite{Qin2018VINS-Mono} exhibited the best performance for their underwater datasets.
Singh \etal~\cite{singhOnlineRefractiveCamera2024} presented a new refractive camera model and an approach for online estimation of refractive index of a medium using stereo vision based on ROVIO.
Later, they extended to a vision-based underwater exploration and inspection system~\cite{Singh2025Aeriel}.
Hidalgo \etal~\cite{Hidalgo2020ORBSLAM2} adapted ORB-SLAM2 for underwater environments by incorporating a robust feature tracking method and an illumination correction algorithm.
Rahman \etal~\cite{Rahman2019SVIn2} proposed the sonar-visual-inertial SLAM system, which integrated a downward-facing pipe-profiling sonar, a stereo camera, and an IMU under a tightly-coupled framework based on OKVIS.
However, none of these methods investigated the incorporation of DVL.

Only a subset of literature incorporates DVL in a visual SLAM pipeline.
Vargas \etal~\cite{Vargas2021Robust} fused a stereo camera, a DVL, and a gyroscope in a loosely-coupled fashion.
Thomas \etal~\cite{Thomas2023Tightly} integrated DVL into a LiDAR-visual-inertial system for surface vehicles that operate on 2D manifolds.
Fusing DVL and depth sensor into tightly-coupled visual-inertial odometry was proposed by~\cite{Zhao2023Tightly} in a filtering-based method.
Huang \etal~\cite{Huang2023Tightly} proposed a visual-DVL fusion method, which integrates the velocity measurements from a DVL into an optimization-based visual odometry pipeline for improved accuracy.
However, the lack of IMU integration compromised the robustness of orientation estimation in challenging visual conditions.
To address this, Xu \etal~\cite{Xu2025AQUA-SLAM} proposed integrating DVL into a tightly-coupled visual-inertial SLAM system, which employs the joint preintegration of IMU and DVL measurements in a graph-based optimization framework.
However, this approach makes fusing other modalities such as depth difficult.
SurfSLAM~\cite{bagorenSurfSLAMSimtoRealUnderwater2026} has recently incorporated global registration of dense depth maps into the TURTLMap~\cite{Song2024TURTLMap} architecture but does not perform bundle adjustment, which limits its methodology to frame-to-frame registration.
To address this problem, this article presents a continuous-time acoustic-visual-inertial odometry system that not only tightly couples these sensors but can seamlessly incorporate additional sensor modalities.

\subsection{Continuous-time Estimation}
\label{subsec:relatedwork-ctestimation}
Within the last decade, continuous-time trajectory estimation has gained significant attention in the state estimation community~\cite{Talbot2025Continuous}.
The main advantage of continuous-time batch estimation is that the trajectory estimate can be sampled at any arbitrary time, making it straightforward to fuse asynchronous measurements and high-frequency sensors without having to increase the number of optimization parameters~\cite{Johnson2024Continuous}.
Two types of continuous-time estimation have become prevalent in the literature: spline-based estimation~\cite{Sommer2015Continuous, Mueggler2018Continuous, Lv2021CLINS} and Gaussian process (GP) regression~\cite{Barfoot2014BatchCT, Anderson2015STEAM, Burnett2025Continuous, Zhang2024GNSSFGO, Wu2023Picking}.

Splines~\cite{Cox1972Numerical} are piecewise polynomial functions that are characterized by a set of control points on the configuration manifold and a set of knot points in time.
These control point locations are optimized such that the resulting trajectory optimally fits the acquired measurements and motion priors~\cite{Furgale2012Continuous}.
Splines have been used for sensor fusion using IMUs with rolling shutter cameras~\cite{PatronPerez2015Spline}, event cameras~\cite{Mueggler2018Continuous}, radar~\cite{Ng2021Continuous}, and LiDAR~\cite{Lv2021CLINS}, and achieved real-time performance.

With GP regression, the robot's states are modeled at a discrete set of estimation times, and the states at any arbitrary time are evaluated using GP interpolation.
The interpolation scheme is derived from models used to represent the motion of the robot such as white-noise-on-acceleration~\cite{Anderson2015STEAM}, white-noise-on-jerk~\cite{Tang2019WNOJ}, and data-driven motion priors~\cite{Wong2020DataDriven}.
GP regression has been used in fusing GNSS~\cite{Zhang2024GNSSFGO}, radar~\cite{Burnett2022Radar}, LiDAR~\cite{Burnett2025Continuous}, and even continuum robot pose estimation~\cite{Lilge2025State}.

A direct comparison between spline and GP-based estimation was done in~\cite{Johnson2024Continuous} by comparing them in a camera and IMU sensor fusion scenario on ${SE(3)}$.
Their results indicate that the two methods achieve similar trajectory accuracy if the same motion model and measurements are used.
Solve times are also similar so long as the spline order is chosen such that the degree-of-differentiability matches that of the GP.
Choosing a higher spline order did not improve trajectory accuracy, while requiring significantly higher solve times.

\section{Continuous-time Trajectory Estimation}
\label{sec:continuousestimation}
In this section, continuous-time trajectory estimation using Gaussian processes~\cite{Barfoot2014BatchCT, Anderson2015STEAM} is reviewed.

\subsection{White-noise-on-acceleration Motion Model}
\label{subsec:continuousestimation-wnoa}
The choice of motion prior is important within the continuous-time estimation framework. Consider the following time-varying stochastic differential equation (SDE),
\begin{subequations}
    \label{eq:global_sde}
    \begin{align}
        \mbfdot{T}(t)      & = \mbf{T}(t) \mbs{\varpi}(t)^{\wedge}, \quad (\cdot)^{\wedge}: \mathbb{R}^{6} \mapsto \mathfrak{se}(3), \\
        \mbsdot{\varpi}(t) & = \mbf{w}'(t), \quad \mbf{w}'(t) \sim \mc{GP}\left(\mbf{0}, \mbc{Q}'\delta(t-t')\right),
    \end{align}
\end{subequations}
where ${\mbf{T}(t) \in SE(3)}$ is the pose, ${\mbs{\varpi}(t) = \bbm \mbs{\omega}^{\trans} & \mbs{\nu}^{\trans} \ebm^{\trans} \in \mathbb{R}^{6}}$ is the velocity resolved in the body frame consisting of an angular ${\mbs{\omega}(t)}$ and linear ${\mbs{\nu}(t)}$ component. A zero-mean, white-noise Gaussian process ${\mbf{w}'(t)}$ is characterized by a symmetric positive-definite power-spectral density matrix ${\mbc{Q}'}$ and ${\delta\left(\cdot\right)}$ denotes the Dirac delta function.
This model is referred to as white-noise-on-acceleration (WNOA) due to white noise being injected on the body-centric acceleration ${\mbsdot{\varpi}(t)}$~\cite{Barfoot2014BatchCT}, which is appropriate for applications with approximately constant-velocity motion~\cite{Burnett2025Continuous}.
\subsection{Local Gaussian Process on $SE(3)$}
\label{subsec:continuousestimation-localgp}

\begin{figure}[t]
    \centering
    \includegraphics[width=0.95\linewidth]{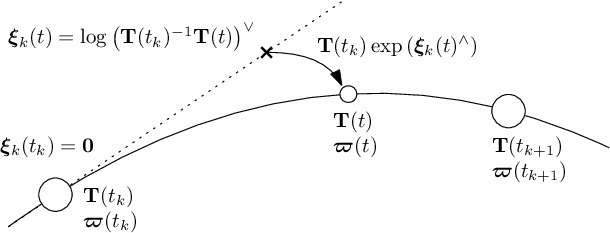}
    \caption{The relationship between the local variable ${\mbs{\xi}_{k}(t)}$, which is in the tangent space of the pose at time ${t_{k}}$, and the global pose ${\mbf{T}(t)}$.}
    \label{fig:gp}
\end{figure}

Anderson \etal~\cite{Anderson2015STEAM} defines a local linear time-invariant (LTI) SDE between each pair of estimation times ${t_{k}}$ and ${t_{k+1}}$ for ${k \in [0, K-1]}$ first by defining a local pose variable as
\begin{align}
    \label{eq:local_pose}
    \mbs{\xi}_{k}(t) & = \Log\left(\mbf{T}(t_{k})\inv\mbf{T}(t)\right), \quad t \in [t_{k}, t_{k+1}],
\end{align}
where ${\Log(\cdot)} = \log((\cdot)^{\vee})$ is the matrix logarithm map~\cite{Sola2018Micro}.
The relationship between local pose ${\mbs{\xi}_{k}(t)}$ and the global pose ${\mbf{T}(t)}$ is illustrated in Fig.~\ref{fig:gp}.
The kinematics of the local variable is given in~\cite[\S 8.2.2]{Barfoot2023State} as
\begin{align}
    \label{eq:local_kinematics}
    \mbsdot{\xi}_{k}(t) & = \mbf{J}_{r}\inv(\mbs{\xi}_{k}(t))\mbs{\varpi}(t),
\end{align}
where $\mbf{J}_{r}$ is the right Jacobian of $SE(3)$~\cite[\S 8.1.5]{Barfoot2023State}.
Note that the right Jacobian is used here due to the convention of the pose.
Based on the WNOA assumption, a sequence of local LTI SDEs can be constructed as
\begin{align}
    \label{eq:local_sde}
    \mbsddot{\xi}_{k}(t) = \mbf{w}(t), \quad \mbf{w}(t) \sim \mc{GP}\left(\mbf{0}, \mbc{Q}\delta(t-t_k)\right),
\end{align}
where hyperparameter ${\mbc{Q} \in \mathbb{R}^{6 \times 6}}$ is the symmetric power spectral density of the white noise process and can be tuned for different applications~\cite{Wong2020DataDriven}.

The local Markovian state variables are defined as
\begin{align}
    \label{eq:local_markovian_state}
    \mbs{\gamma}_{k}(t) = \begin{bmatrix}
                              \mbs{\xi}_{k}(t) \\
                              \mbsdot{\xi}_{k}(t)
                          \end{bmatrix} \in \mathbb{R}^{12}.
\end{align}
The local LTI SDE that can be written as
\begin{align}
    \label{eq:local_lti_sde}
    \mbsdot{\gamma}_{k}(t) & = \begin{bmatrix}
                                   \mbf{0} & \mbf{1} \\
                                   \mbf{0} & \mbf{0}
                               \end{bmatrix}
    \mbs{\gamma}_{k}(t) + \begin{bmatrix}
                              \mbf{0} \\
                              \mbf{1}
                          \end{bmatrix} \mbf{w}(t),
\end{align}
where $\mbf{1}$ is the identity matrix. This can then be integrated to yield the local GP,
\begin{equation}
    \label{eq:local_gp}
    \begin{split}
        \mbs{\gamma}_{k}(t) \sim \mc{GP}\Big(&\mbs{\Phi}(t, t_{k})\mbscheck{\gamma}_{k}(t_{k}), \\
        & \mbs{\Phi}(t, t_{k})\mbfcheck{P}(t_{k})\mbs{\Phi}(t, t_{k})^{\trans} + \mbf{Q}(t_{k})\Big),
    \end{split}
\end{equation}
where ${\mbscheck{\gamma}_{k}(t) = \expect{\mbs{\gamma}_{k}(t)}}$, ${\mbfcheck{P}(t_{k})}$ is the initial covariance,
\begin{align}
    \mbs{\Phi}(t, t_k) & = \begin{bmatrix}
                               \mbf{1} & (t - t_k)\mbf{1} \\
                               \mbf{0} & \mbf{1}
                           \end{bmatrix}, \quad t \geq t_k,
\end{align}
with the corresponding covariance given as
\begin{align}
    \label{eq:local_covariance}
    \mbf{Q}(t, t_k) & = \begin{bmatrix}
                            \frac{1}{3}(t - t_k)^{3}\mbc{Q} & \frac{1}{2}(t - t_k)^{2}\mbc{Q} \\
                            \frac{1}{2}(t - t_k)^{2}\mbc{Q} & (t - t_k)\mbc{Q}
                        \end{bmatrix}.
\end{align}
The exact sparsity (block-tridiagonality) of the inverse kernel matrix is exploited to enable efficient batch estimation that scales linearly with the number of estimation times~\cite{Anderson2015STEAM}.

\subsection{Gaussian-process Motion Prior}
\label{subsec:continuousestimation-gpprior}
The WNOA residual between two consecutive estimation times ${t_{k}}$ and ${t_{k+1}}$ can be formulated as
\begin{align}
    \label{eq:gp_residual_unsimp}
    \begin{split}
        \mbf{e}^{GP}_{kk+1} & = \mbs{\gamma}_{k}(t_{k+1}) - \mbscheck{\gamma}_{k}(t_{k+1}) \\ & - \mbs{\Phi}(t_{k+1}, t_{k})(\mbs{\gamma}_{k}(t_{k}) - \mbscheck{\gamma}_{k}(t_{k})),
    \end{split}
\end{align}
With \eqref{eq:local_gp}, the means in \eqref{eq:gp_residual_unsimp} cancel out, and the residual simplifies to
\begin{align}
    \label{eq:gp_residual}
    \mbf{e}^{GP}_{kk+1} & = \mbs{\gamma}_{k}(t_{k+1}) - \mbs{\Phi}(t_{k+1}, t_{k})\mbs{\gamma}_{k}(t_{k}).
\end{align}
Substituting \eqref{eq:local_pose} and \eqref{eq:local_kinematics} into \eqref{eq:local_markovian_state} yields
\begin{align}
    \mbs{\gamma}_{k}(t) = \begin{bmatrix}
                              \mbs{\xi}_{k}(t) \\
                              \mbsdot{\xi}_{k}(t)
                          \end{bmatrix} = \begin{bmatrix}
                                              \Log\left(\mbf{T}(t_{k})\inv\mbf{T}(t)\right) \\
                                              \mbf{J}_{r}\inv\left(\Log\left(\mbf{T}(t_{k})\inv\mbf{T}(t)\right)\right)\mbs{\varpi}(t)
                                          \end{bmatrix}.
\end{align}
\begin{figure}[t]
    \centering
    \includegraphics[width=\linewidth, clip=true, trim={0cm, 5cm, 7cm, 4cm}]{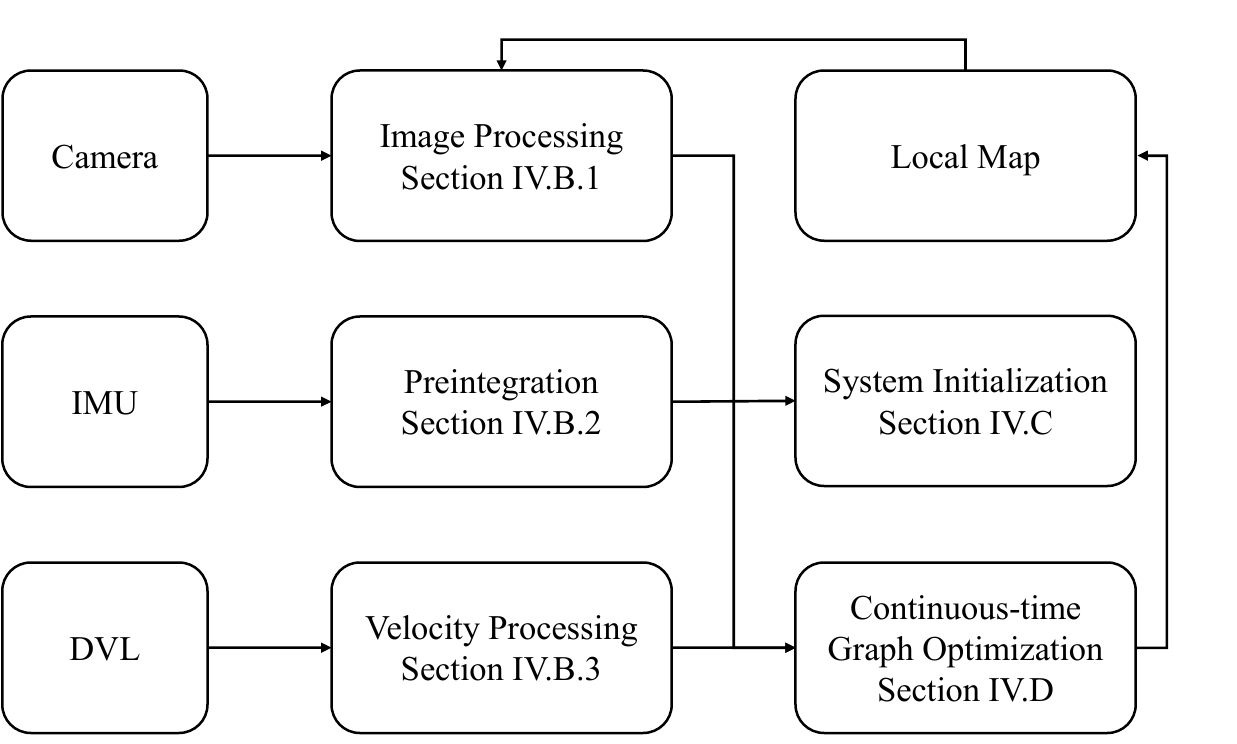}
    \caption{Overall system flowchart.}
    \label{fig:flowchart}
\end{figure}
Thus, the residual can be written as
\begin{align}
    \label{eq:gp_prior_residual}
    \mbf{e}^{GP}_{kk+1} & =
    \begin{bmatrix}
        \Log\left(\mbf{T}_{k}\inv\mbf{T}_{k+1}\right) - \left(t_{k+1} - t_{k}\right)\mbs{\varpi}_{k} \\
        \mbf{J}_{r}\inv\left(\Log\left(\mbf{T}_{k}\inv\mbf{T}_{k+1}\right)\right)\mbs{\varpi}_{k+1} - \mbs{\varpi}_{k}
    \end{bmatrix},
\end{align}
where ${\mbf{T}_{k} = \mbf{T}(t_{k})}$.
The role of the motion prior is to penalize the state estimates from deviating from a constant velocity.

\subsection{Gaussian-process Interpolation}
\label{subsec:continuousestimation-gpinterpolation}
One of the key advantages of continuous-time trajectory estimation is the ability to query the trajectory at any arbitrary time ${t \in [t_{k}, t_{k+1})}$ using GP interpolation.
The interpolation of the local state variable ${\mbs{\gamma}_{k}(t)}$~\cite{Barfoot2014BatchCT} is,
\begin{align}
    \hspace{-7pt}
    \mbshat{\gamma}_{k}(t) & = \mbscheck{\gamma}_{k}(t) +
    \begin{bmatrix}
        \mbs{\Lambda}(t) & \mbs{\Psi}(t)
    \end{bmatrix}
    \begin{bmatrix}
        \mbshat{\gamma}_{k}(t_{k}) - \mbscheck{\gamma}_{k}(t_{k}) \\
        \mbshat{\gamma}_{k}(t_{k+1}) - \mbscheck{\gamma}_{k}(t_{k+1})
    \end{bmatrix},
\end{align}
where ${\mbs{\Lambda}(t)}$ and ${\mbs{\Psi}(t)}$ are
\begin{align}
    \mbs{\Lambda}(t) & = \mbs{\Phi}(t, t_{k}) - \mbs{\Psi}(t)\mbs{\Phi}(t_{k+1}, t_{k}),               \\
    \mbs{\Psi}(t)    & = \mbf{Q}(t, t_{k}) \mbs{\Phi}(t_{k+1}, t)^{\trans}\mbf{Q}(t_{k+1}, t_{k})\inv.
\end{align}
The posterior interpolation can be rewritten using the mean of the local Markovian state variable \eqref{eq:local_gp} as
\begin{align}
    \label{eq:local-gp-interpolation}
    \mbshat{\gamma}_{k}(t) & = \mbs{\Lambda}(t)\mbshat{\gamma}_{k}(t_{k}) + \mbs{\Psi}(t)\mbshat{\gamma}_{k}(t_{k+1}),
\end{align}
which can be decomposed into
\begin{subequations}
    \begin{align}
        \mbshat{\gamma}^{1}_{k}(t) & = \mbshat{\xi}_{k}(t) = \mbs{\Lambda}^{1}(t)\mbshat{\gamma}_{k}(t_{k}) + \mbs{\Psi}^{1}(t)\mbshat{\gamma}_{k}(t_{k+1}),       \\
        \mbshat{\gamma}^{2}_{k}(t) & = \dot{\mbshat{\xi}}_{k}(t) = \mbs{\Lambda}^{2}(t)\mbshat{\gamma}_{k}(t_{k}) + \mbs{\Psi}^{2}(t)\mbshat{\gamma}_{k}(t_{k+1}).
    \end{align}
\end{subequations}
The interpolated local variables are then transformed to the global variables ${\mbf{T}}$ and ${\mbs{\varpi}}$ as
\begin{align}
    \label{eq:gp_interpolation}
    \mbfhat{T}(t) = \mbfhat{T}_{k}\Exp\left(\mbshat{\gamma}^{1}_{k}(t)\right), \quad \mbshat{\varpi}(t) = \mbf{J}_{r}\left(\mbshat{\gamma}^{1}_{k}(t)\right)\mbshat{\gamma}^{2}_{k}(t),
\end{align}
where ${\Exp(\cdot)=\exp((\cdot)^{\wedge})}$ is the matrix exponential map~\cite{Sola2018Micro}.
\section{Continuous-time DVL-Inertial-Visual Odometry System}
\label{sec:divo}
In this section, the proposed DVL-inertial-visual odometry (DIVO) system is presented, starting with the state definition, followed by the frontend processing, system initialization, and backend optimization.
Finally, implementation details conclude this section. An overall system flowchart is shown in Fig.~\ref{fig:flowchart}.

\begin{figure}
    \centering
    \includegraphics[width=\linewidth, clip=true, trim={0cm, 2.5cm, 7.8cm, 2.5cm}]{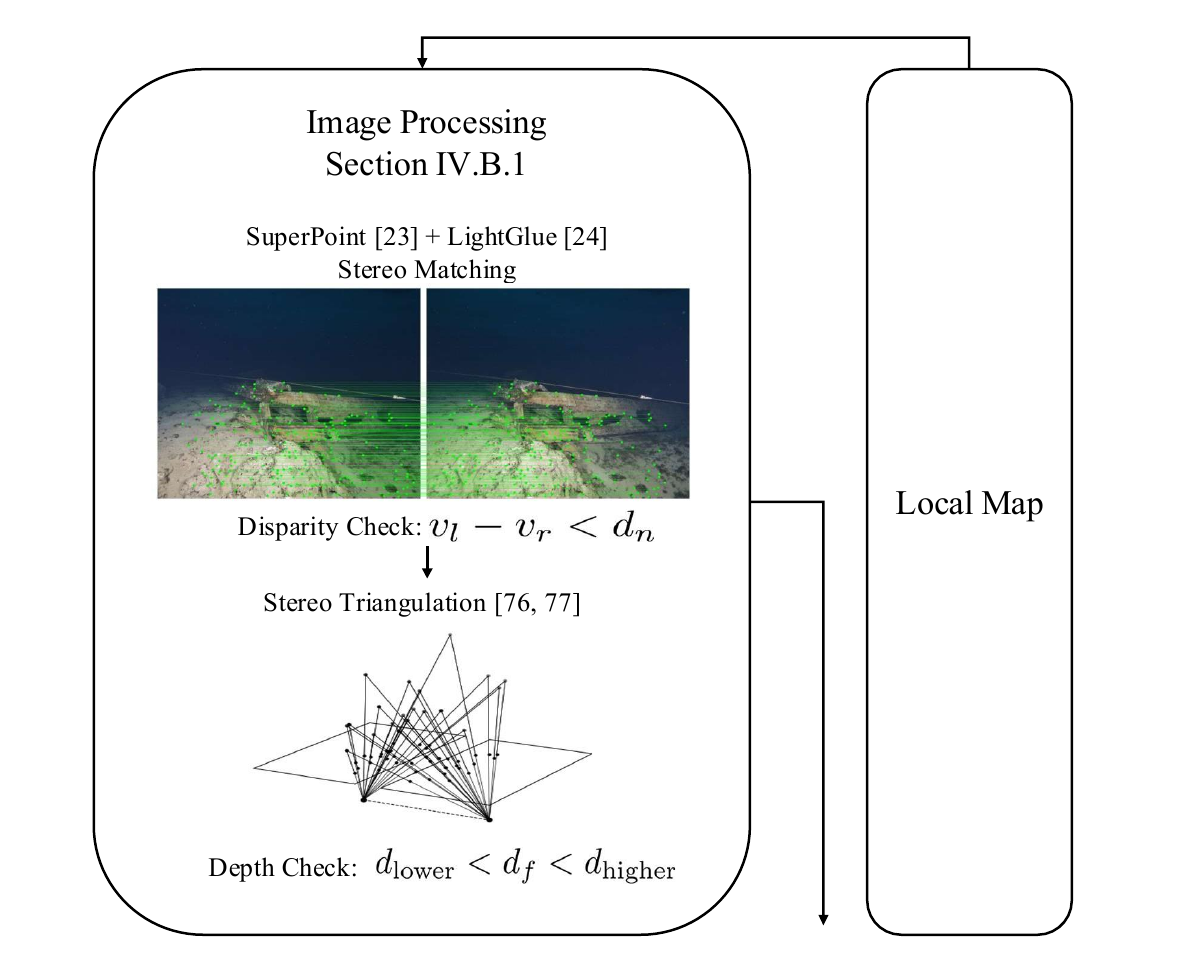}
    \caption{Feature tracking pipeline for the proposed learned frontend.}
    \label{fig:feature_tracking}
\end{figure}
\vspace{0.1cm}
%

\subsection{Notation and State Definition}
\label{subsec:divo-state}
\begin{figure}[t]
    \centering
    \includegraphics[width=0.9\linewidth, clip=true, trim={4cm, 3cm, 1cm, 2cm}]{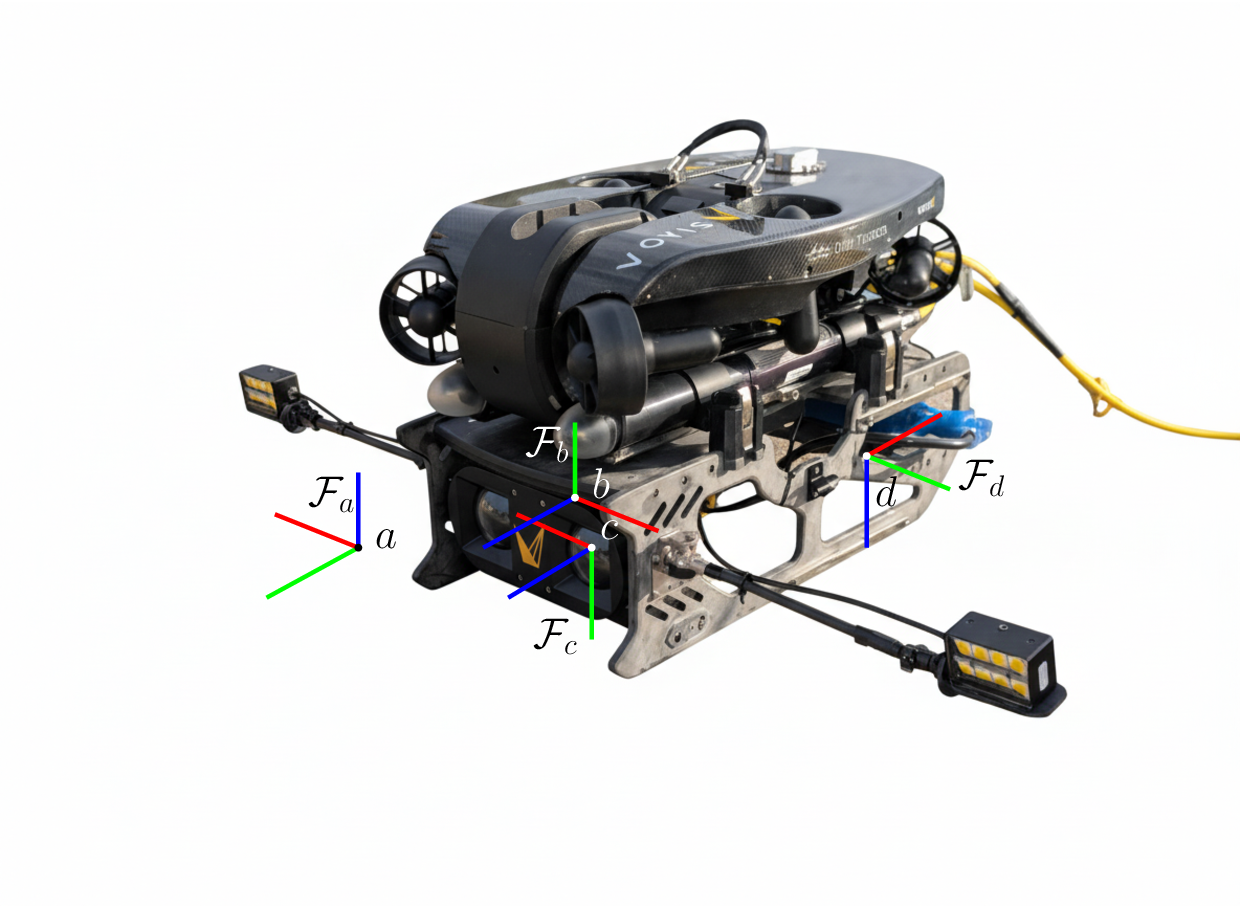}
    \caption{Coordinate frames of the local tangent frame ${\mc{F}_{a}}$, IMU frame ${\mc{F}_{b}}$, camera frame ${\mc{F}_{c}}$, and DVL frame ${\mc{F}_{d}}$.}
    \label{fig:frames}
\end{figure}
The local tangent~\cite{Groves2013Principles}, IMU, left camera, and DVL frames are denoted as ${\mc{F}_{a}}$, ${\mc{F}_{b}}$, ${\mc{F}_{c}}$, and ${\mc{F}_{d}}$, respectively.
The $z$-axis of the local tangent frame is aligned with the gravity direction.
The IMU frame is referred to as the body frame hereafter.
The locations of the local tangent frame datum, IMU, camera, and DVL are denoted as ${a}$, ${b}$, ${c}$, and ${d}$, respectively, as shown in Fig.~\ref{fig:frames}.
The direction cosine matrix (DCM) $\mbf{C}_{ab} \in SO(3)$ describes the change of basis between the body frame $\mathcal{F}_b$ and $\mathcal{F}_a$.
The location of point ${b}$ relative to point ${a}$ resolved in frame ${\mc{F}_{a}}$ is denoted as ${\mbf{r}^{ba}_{a}}$, where superscripts represent the points, and the subscript represents the frame that the physical vector is resolved in.
The DCM ${\mbf{C}_{ab}}$ and position ${\mbf{r}^{ba}_{a}}$ can be concisely written in terms of the transformation matrix
\begin{align}
    \mbf{T}_{ab} = \begin{bmatrix}
                       \mbf{C}_{ab} & \mbf{r}^{ba}_{a} \\
                       \mbf{0}      & 1
                   \end{bmatrix} \in SE(3).
\end{align}
The time derivative of ${\mbf{r}^{ba}_{a}}$ with respect to frame ${\mc{F}_{a}}$ is denoted as ${\mbfdot{r}^{ba}_{a} = \mbf{v}^{ba/a}_{a}}$, where the slash notation in the superscript serves to denote that the time derivative is with respect to frame ${\mc{F}_{a}}$. An extended pose matrix that includes the orientation, position, and linear velocity~\cite{Barrau2017IEKF}
\begin{align}
    \mbs{T}_{ab} = \begin{bmatrix}
                       \mbf{C}_{ab} & \mbf{v}^{ba/a}_{a} & \mbf{r}^{ba}_{a} \\
                       \mbf{0}      & 1                  & 0                \\
                       \mbf{0}      & 0                  & 1
                   \end{bmatrix} \in SE_2(3).
\end{align}

The state at each camera frame ${k}$ is defined as
\begin{align}
    \label{eq:state}
    \mc{X}_{k} = (\mbs{T}_{ab_{k}}, \mbf{b}_{k}, \mbs{\omega}^{ba}_{b_{k}}, \{\mbf{r}^{\ell_{m}b_{k}}_{b_{k}}\}_{m=0}^{M_{k}}),
\end{align}
where ${\mbf{b}_{k}}$ is the IMU bias consisted of the gyroscope and accelerometer biases, and ${\mbs{\omega}^{ba}_{b_{k}}}$ is the angular velocity of frame ${\mc{F}_{b}}$ relative to frame ${\mc{F}_{a}}$ resolved in frame ${\mc{F}_{b}}$.
In~\cite{Burnett2025Continuous}, a GP motion prior is utilized and the estimated state consists of ${(\mbf{T} \in SE(3), \mbs{\varpi} \in \mathbb{R}^{6})}$ where ${\mbs{\varpi} = \bbm \mbs{\omega}^{\trans} & \mbs{\nu}^{\trans} \ebm^{\trans}}$ is the body-centric velocity. As such, gyroscope measurements are treated as a direct measurement of the state, while accelerometer measurements are preintegrated to form relative velocity constraints. However, in this work, inspiration is drawn from~\cite{Zhang2024GNSSFGO}, where the GP motion prior is utilized alongside IMU preintegration. Thus, the state is instead defined as in~\eqref{eq:state} to handle compact IMU preintegration~\cite{Brossard2022Associating} and a GP motion prior.
The set of all states in a sliding window size of $n$ is then defined as
\begin{align}
    \label{eq:state_sliding_window}
    \mc{X} = \left\{\mc{X}_{k-n}, \ldots, \mc{X}_{k}\right\}.
\end{align}
At every ${k}$-th image, ${M_{k}}$ visual features that are first observed at that image are included in the state.
The 3D position of the $m$-th feature anchored at the $k$-th image is denoted as ${\mbf{r}^{\ell_{m}b_{k}}_{b_{k}}}$.

Additionally, the ${\oplus}$ and ${\ominus}$ operators, which will be used in the forthcoming derivations, denote the left plus and minus operators applied to Lie group elements as defined in~\cite{Sola2018Micro}. The derivative of ${f(\mc{Z})}$ with respect to a Lie group element~${\mc{Z}}$ is then defined as~\cite{Sola2018Micro}
\begin{align}
    \frac{Df(\mc{Z})}{D\mc{Z}} = \lim_{\mbs{\tau}\to\mbf{0}} \frac{f\left(\mc{Z}\oplus\mbs{\tau}\right)\ominus f(\mc{Z})}{\mbs{\tau}}.
\end{align}

\subsection{Frontend Processing}
\label{subsec:divo-frontend}
The frontend processing consists of three parts: image processing, IMU preintegration, and DVL processing.
\vspace{0.1cm}

\subsubsection{Processing Image Data}
\label{subsubsec:divo-frontend-image}
The tracking schemes for both GPU-based SuperPoint-LightGlue and CPU-based KLT tracking are presented.
The implementation details for both are presented in Section~\ref{subsec:divo-implementation}.
\vspace{0.1cm}
\subsubsection*{SuperPoint-LightGlue (SL) Tracking}
At every new image, up to 512 SuperPoint~\cite{detoneSuperPointSelfSupervisedInterest2018} features are detected.
Their corresponding stereo features are found by left-to-right matching, and the corresponding stereo feature pairs with a disparity above a noise threshold ${d_{n}}$ are rejected.
The initial feature instances are triangulated using LOST triangulation ~\cite{Henry2023LOST}, and each subsequent instance is triangulated using DLT~\cite{abdel-azizDirectLinearTransformation2015} to get their 3D positions relative to the body.
The feature instance is rejected if its depth does not lie within the upper and lower depth bounds.
If the disparity and depth check pass, the feature is added to the feature manager.
LightGlue~\cite{lindenbergerLightGlueLocalFeature2023} is then employed to match all the features belonging to camera frame ${\mathcal{F}_{c_{k-n}}}$ to current frame ${\mathcal{F}_{c_{k}}}$.
The frame-to-frame match is rejected if the number of matched features does not lie above some threshold ${m_{n}}$. The proposed pipeline is visualized in Fig.~\ref{fig:feature_tracking}.

\subsubsection*{KLT Tracking}
The tracking and mapping scheme is adopted from~\cite{Qin2019General}.
At every image, Shi-Tomasi features~\cite{Shi1994Good} are detected and tracked using the Kanade-Lucas-Tomasi (KLT) tracker~\cite{Lucas1981KLT}.
First, the existing features from the previous image are tracked to the current image.
Falsely tracked features, also known as outliers, are rejected using the essential matrix computed using the five-point algorithm~\cite{Nister2004Five} within a RANSAC scheme~\cite{Fischler1981RANSAC}.
Then, new features are detected in the current image until a maximum number of features is reached.
For newly detected features, corresponding stereo features are found using the KLT tracker.
Stereo outliers are rejected using the same method.
Lastly, the stereo matches are triangulated using LOST triangulation~\cite{Henry2023LOST} to get their 3D positions relative to the body at which they are first observed.
\vspace{0.1cm}

\subsubsection{Processing IMU Data}
\label{subsubsec:divo-frontend-imu}
Consider the continuous-time kinematics of a vehicle on the rotating Earth in a compact form~\cite{Shalaby2024Multi},
\begin{gather}
    \label{eq:imu_kinematics}
    \mbsdot{T}_{ab}(t) = \mbf{G}\mbs{T}_{ab}(t) + \mbs{T}_{ab}(t)\mbf{U}(t), \\
    \hspace{-8pt}
    \mbf{G} = \begin{bmatrix}
        -\left(\mbs{\omega}^{ai}_{a}\right)^{\times} & \mbf{g}_{a} & \mbf{0} \\
        \mbf{0}                                      & 0           & -1      \\
        \mbf{0}                                      & 0           & 0
    \end{bmatrix}, \,\,
    \mbf{U} = \begin{bmatrix}
        \left(\mbs{\omega}^{bi}_{b}\right)^{\times} & \mbf{f}_{b} & \mbf{0} \\
        \mbf{0}                                     & 0           & 1       \\
        \mbf{0}                                     & 0           & 0
    \end{bmatrix},
\end{gather}
where ${(\cdot)^{\times}: \mathbb{R}^{3} \mapsto \mf{so}(3)}$, ${\mbs{\omega}^{ai}_{a}}$ is the angular velocity of the local tangent frame $\mc{F}_{a}$ relative to the Earth-Centered Inertial frame ${\mc{F}_{i}}$ resolved in $\mathcal{F}_a$, ${\mbf{g}_{a}}$ is the local gravity resolved in the local tangent frame ${\mc{F}_{a}}$, ${\mbs{\omega}^{bi}_{b}}$ is the angular velocity of ${\mc{F}_{b}}$, relative to ${\mc{F}_{i}}$, resolved in $\mc{F}_{b}$, measured by the gyroscope, and ${\mbf{f}_{b}}$ is the specific force measured by the accelerometer resolved in ${\mc{F}_{b}}$.
In this paper, the effect of Earth's rotation is neglected, meaning ${\mbs{\omega}^{ai}_{a} = \mbf{0}}$.

The compact continuous-time kinematics has a form of a \emph{differential Sylvester equation}~\cite{Shalaby2024Multi, Behr2019Solution}.
Assuming the input matrix ${\mbf{U}}$ at time ${t_{k-1}}$ is constant over the integration period ${\Delta{t} = t_{k} - t_{k-1}}$, the solution to \eqref{eq:imu_kinematics} with the initial condition ${\mbs{T}_{ab_{k-1}}}$ is given by
\begin{subequations}
    \label{eq:imu_kinematics_discrete}
    \begin{align}
        \mbs{T}_{ab_{k}} & = \exp\left(\Delta{t}\mbf{G}\right)\mbs{T}_{ab_{k-1}}\exp\left(\Delta{t}\mbf{U}(t_{k-1})\right) \\
                         & = \mbf{G}_{k-1}\mbs{T}_{ab_{k-1}}\mbf{U}_{k-1}.
    \end{align}
\end{subequations}

The gyroscope and accelerometer measurements are modeled as
\begin{subequations}
    \begin{align}
        \mbstilde{\omega}^{bi}_{b} & = \mbs{\omega}^{bi}_{b} + \mbf{b}^{g} + \mbs{\eta}^{g},                                                \\
        \mbftilde{f}_{b}           & = \mbf{C}_{ab}^{\trans}\left(\mbf{a}^{bo/i/i}_{a} - \mbf{g}_{a}\right) + \mbf{b}^{a} + \mbs{\eta}^{a},
    \end{align}
\end{subequations}
where~${\mbs{\eta}^{g} \sim \mc{N}(\mbf{0}, \mbc{Q}^{g}\delta(t-\tau))}$~and~${\mbs{\eta}^{a} \sim \mc{N}(\mbf{0}, \mbc{Q}^{a}\delta(t-\tau))}$ are zero-mean white Gaussian process noise, and ${\mbf{b}^{g}}$ and ${\mbf{b}^{a}}$ are random walk biases for the gyroscope and accelerometer biases, respectively.
Substituting the measurements into the input matrix ${\mbf{U}}$ yields
\begin{align}
    \mbftilde{U} = \begin{bmatrix}
                       \left(\mbstilde{\omega}^{bi}_{b} - \mbf{b}^{g} - \mbs{\eta}^{g}\right)^{\times} & \mbftilde{f}_{b} - \mbf{b}^{a} - \mbs{\eta}^{a} & \mbf{0} \\
                       \mbf{0}                                                                         & 0                                               & 1       \\
                       \mbf{0}                                                                         & 0                                               & 0
                   \end{bmatrix}.
\end{align}

The compact form \eqref{eq:imu_kinematics_discrete} can be used to easily propagate multiple states from some arbitrary time ${t_{i}}$ to ${t_{j}}$ by
\begin{subequations}
    \label{eq:imu_propagation}
    \begin{align}
        \mbs{T}_{j} & = \left(\prod_{k=i}^{j-1} \mbf{G}_{k}\right) \mbs{T}_{i} \left(\prod_{k=i}^{j-1} \mbftilde{U}_{k}\right) \\
                    & = \mbf{G}_{ij} \mbs{T}_{i} \mbftilde{U}_{ij},
    \end{align}
\end{subequations}
where ${\mbf{G}_{ij}}$ is a constant gravity matrix, and ${\mbf{U}_{ij}}$ is the preintegrated IMU measurement~\cite{Forster2017Preintegration, Brossard2022Associating}.
Note that the subscripts in \eqref{eq:imu_kinematics_discrete} have been replaced with subscripts indicating time in \eqref{eq:imu_propagation}.
Because preintegration can be computed incrementally as
\begin{align}
    \label{eq:rmi}
    \mbftilde{U}_{ij} & = \mbftilde{U}_{ij-1}\exp\left(\Delta{t}\mbftilde{U}(t_{j-1})\right) = \begin{bmatrix}
                                                                                                   \Delta\mbf{C}_{ij} & \Delta\mbf{v}_{ij} & \Delta\mbf{r}_{ij} \\
                                                                                                   \mbf{0}            & 1                  & \Delta{t}_{ij}     \\
                                                                                                   \mbf{0}            & 0                  & 1
                                                                                               \end{bmatrix},
\end{align}
the kinematics of the preintegrated IMU measurement can also be formulated in continuous time as
\begin{align}
    \label{eq:rmi_kinematics}
    \mbsdot{\Xi}(t) = \mbs{\Xi}(t)\mbftilde{U}(t),
\end{align}
where ${\mbs{\Xi}(t)}$ is the preintegrated IMU measurement up to time ${t}$, and ${\mbftilde{U}(t)}$ is the input matrix.
The error-state dynamics of the preintegrated IMU measurement can then be derived using linearization of \eqref{eq:rmi_kinematics} as
\begin{align}
    \delta\mbsdot{\zeta} & = \mbf{A}\mbsdel{\zeta} + \mbf{L}\mbsdel{\eta},
\end{align}
where ${\Exp(\mbsdel{\zeta}) = \mbsdel{\Xi}}$, and ${\mbf{A} = D\mbsdot{\Xi}/D\mbs{\Xi}}$ and ${\mbf{L} = D\mbs{\Xi}/D\mbs{\eta}}$ are the Jacobians of the continuous-time preintegrated IMU measurement model with respect to the preintegrated IMU measurement and the IMU noise, respectively.
This allows the covariance to be propagated in continuous time.
\vspace{0.1cm}

\subsubsection{Processing DVL Data}
\label{subsubsec:divo-frontend-dvl}
A DVL typically consists of four transducers.
Each transducer is oriented towards the seafloor at a known angle, sending out an oscillating acoustic signal~\cite{Rudolph2012Doppler}
\begin{align}
    s(t) = A_{0}(t)\sin\left(2\pi f_{0}t + \phi_{0}(t)\right),
\end{align}
where ${A_{0}}$ is the amplitude, ${f_{0}}$ is the output frequency, and ${\phi_{0}(t)}$ is the phase.
The transducer samples the return signal that has the form
\begin{align}
    r(t) = A_{1}(t)\sin\left(2\pi f_{1}t + \phi_{1}(t) + \eta(t)\right),
\end{align}
where ${\eta(t)}$ is zero-mean white Gaussian process noise.
The velocity in the direction of the transducer can then be found by rearranging the Doppler equation,
\begin{align}
    v = \f{2c(f_{1} - f_{0})}{f_{0}},
\end{align}
where ${c}$ is the speed of sound in water~\cite{Rudolph2012Doppler}.

\begin{figure}[t]
    \centering
    \includegraphics[width=0.9\linewidth]{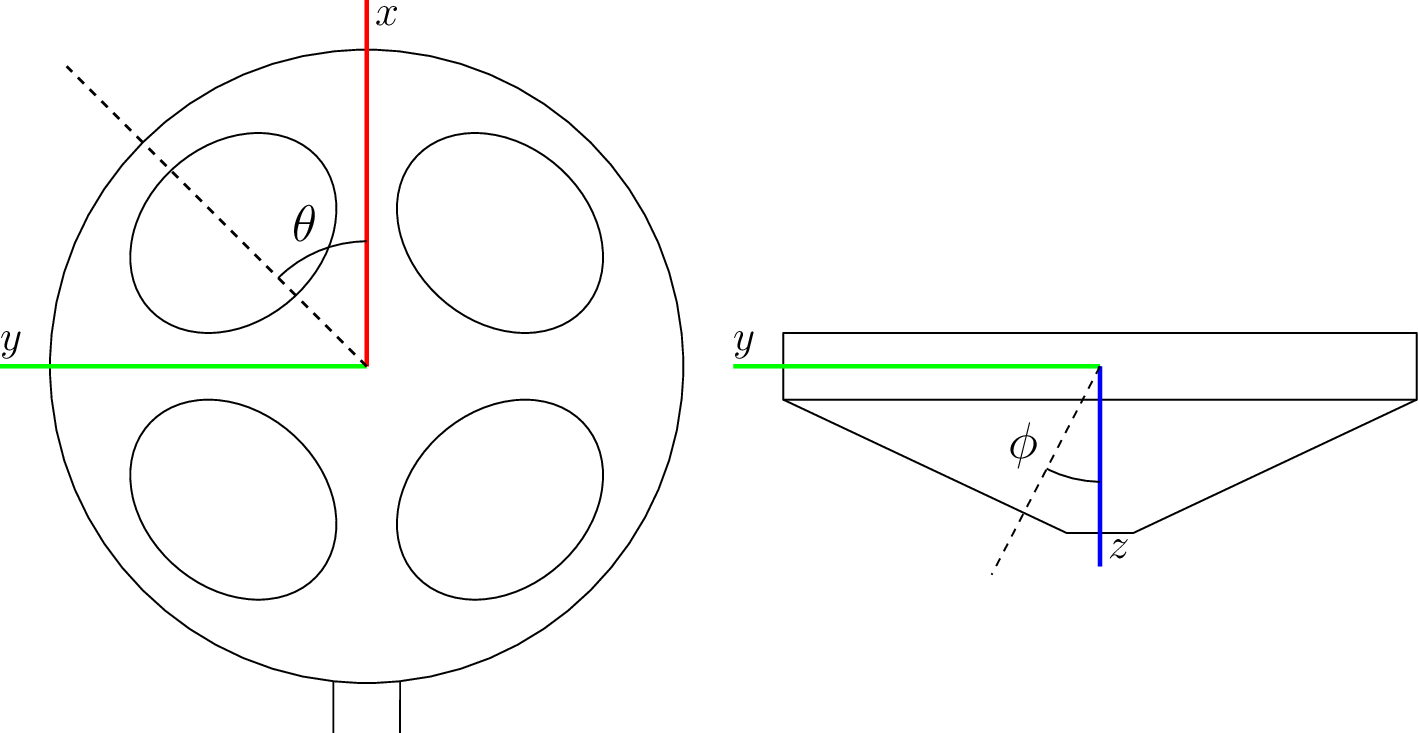}
    \caption{
        DVL transducer configuration.
        Each transducer is oriented at an azimuth ${\theta}$ and elevation ${\phi}$ with respect to the DVL frame.
    }
    \label{fig:dvl}
\end{figure}
Each beam angle is parametrized by the azimuth ${\theta}$ and the elevation angle ${\phi}$ as shown in Fig.~\ref{fig:dvl}.
The direction cosine matrix from the DVL frame to the ${j}$-th transducer frame can be constructed from a 3-2-1 Euler angle sequence as
\begin{align}
    \mbf{C}_{d\tau_{j}} = \mbf{C}_{3}(\theta_{j})\mbf{C}_{2}(\phi_{j})\mbf{C}_{1}(0).
\end{align}
%
The relationship between the ${j}$-th transducer velocity and the DVL velocity is described using the Transport Theorem,
\begin{align}
    v_{j} & = \mbf{d}_{3}^{\trans}\mbf{C}_{d\tau_{j}}^{\trans}\left(\mbf{v}^{da/a}_{d} - \left(\mbs{\omega}^{\tau_{j}i}_{d} + \mbf{C}_{ad}^{\trans}\mbs{\omega}^{ai}_{a}\right)^{\times}\mbf{r}^{\tau_{j}v}_{d}\right),
    \label{eq:dvl_beam_velocity}
\end{align}
where ${\mbf{d}_{3} = [0\;0\;1]^{\trans}}$, ${\mc{F}_{\tau_{j}}}$ is the frame of the ${j}$-th transducer where its ${z}$-axis is aligned with the acoustic beam direction, ${(\mbf{C}_{d\tau_{j}}, \mbf{r}^{\tau_{j}v}_{d})}$ is the extrinsic calibration between the DVL and transducer frames, and ${\mbs{\omega}^{\tau_{j}i}_{d}}$ is the angular velocity of the transducer frame ${\mathcal{F}_{\tau_j}}$ relative to the ECI frame ${\mathcal{F}_i}$ resolved in the DVL frame ${\mathcal{F}_d}$.
Neglecting the Earth's rotation effect and the small Coriolis effect caused by the small lever arm between the DVL and transducer, \eqref{eq:dvl_beam_velocity} becomes
\begin{subequations}
    \begin{align}
        v_{j} & = \mbf{d}_{3}^{\trans}\mbf{C}_{d\tau_{j}}^{\trans}\mbf{v}^{da/a}_{d},
    \end{align}
\end{subequations}
where ${\mbf{d}_{3}^{\trans}\mbf{C}_{d\tau_{j}}^{\trans} = \mbf{e}_{j}^{\trans}}$, and $\mbf{e}_{j}$ is a canonical basis vector.
Finally, the DVL velocity is obtained by solving the linear least-squares problem~\cite{Rudolph2012Doppler},
\begin{align}
    \label{eq:dvl_velocity}
    \begin{bmatrix}
        \mbf{e}_{1} & \mbf{e}_{2} & \mbf{e}_{3} & \mbf{e}_{4}
    \end{bmatrix}^{\trans}
    \mbf{v}^{da/a}_{d} & = \begin{bmatrix}
                               v_1 & v_2 & v_3 & v_4
                           \end{bmatrix}^{\trans}.
\end{align}
Failure to obtain a valid DVL velocity measurement can occur when more than one transducer loses bottom lock.
Then, that velocity is considered invalid.
Assuming the measurement noise of each transducer is independent and identically distributed ${\tilde{v}_{j} = v_{j} + \eta_{j}}$ such that  ${\eta \sim \mc{N}(0, \sigma_{j}^{2})}$, the noise of the DVL velocity measurement via linear transformation is also assumed Gaussian.

\subsection{System Initialization}
\begin{figure}[t]
    \centering
    \includegraphics[width=0.9\linewidth]{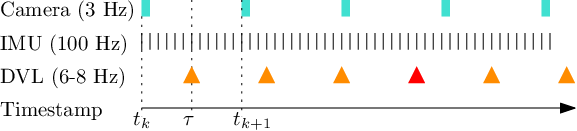}
    \caption{
        Asynchronous sensor measurements in the window used for state estimation.
        States are estimated at every camera frame.
        High frequency IMU measurements are preintegrated.
        The asynchronous DVL measurement at time ${\tau}$ is some time between the two camera frames at ${t_{k}}$ and ${t_{k+1}}$.
        Invalid DVL measurement shown in red is not used in the optimization.
    }
    \label{fig:window}
\end{figure}
\label{subsec:divo-initialization}
The initialization process is necessary to provide a good initial guess of \eqref{eq:state} for the nonlinear optimization. The system initialization is based on the method proposed in~\cite{Qin2018VINS-Mono}.
The process is divided into three parts: vision-only pose estimation, gyroscope bias estimation, and gravity and velocity estimation.
\vspace{0.1cm}

\subsubsection{Vision-only Pose Estimation}
\label{subsubsec:divo-initialization-vo}
Using the temporally tracked features that are triangulated in the frontend, the pose of each camera frame relative to the initial camera frame, ${\mbf{T}_{c_{0}c_{k}}}$, is estimated using the PnP algorithm~\cite{Lepetit2009EPnP}.
Until a sufficient number of consecutive frames ${K}$ are processed, the poses are refined using motion-only bundle adjustment~\cite{Triggs2000Bundle}.
\vspace{0.1cm}

\subsubsection{Gyroscope Bias Estimation}
\label{subsubsec:divo-initialization-gyro}
Once the poses are estimated, the gyroscope bias is estimated by solving
\begin{align}
    {\mbf{b}^{g}}^{\star} = \argmin_{\mbf{b}^{g}} \sum_{k=1}^{K-1} \Vert \Delta\mbf{C}_{kk+1}(\mbf{b}^{g}) \ominus \mbf{C}_{b_{k+1}b_{k}}\Vert^{2}_{2},
\end{align}
where ${\Delta\mbf{C}_{kk+1}(\mbf{b}^{g})}$ is the rotation component of \eqref{eq:rmi} between two consecutive camera frames ${k}$ and ${k+1}$.
The relative rotation of the body is computed by
\begin{align}
    \mbf{C}_{b_{k+1}b_{k}} = \mbf{C}_{bc}^{\trans}\mbf{C}_{c_{0}c_{k+1}}^{\trans}\mbf{C}_{c_{0}c_{k}}\mbf{C}_{bc},
\end{align}
where ${\mbf{C}_{c_{0}c_{k}}}$ is obtained from the relative vision-only pose estimation and ${\mbf{C}_{bc}}$ is the camera-IMU extrinsics.
\vspace{0.1cm}

\subsubsection{Gravity and Velocity Estimation}
\label{subsubsec:divo-initialization-gravity}
After the gyroscope bias initialization, the gravity and velocity are initialized by solving a linear least-squares problem as proposed in~\cite{Qin2018VINS-Mono}.
For a pair of consecutive camera frames ${i}$ and ${j}$, the linear equation is constructed as
\begin{subequations}
    \begin{gather}
        \mbf{A}_{ij}
        \begin{bmatrix}
            \mbf{v}^{b_{i}a/a}_{b_{i}} \\
            \mbf{v}^{b_{j}a/a}_{b_{j}} \\
            \mbf{g}_{b_{0}}
        \end{bmatrix} =
        \begin{bmatrix}
            \mbf{C}_{b_{0}b_{i}}\Delta\mbf{v}_{ij} \\
            \mbf{C}_{b_{0}b_{i}}\Delta\mbf{r}_{ij} - \left(\mbf{r}^{b_{j}b_{0}}_{b_{0}} - \mbf{r}^{b_{i}b_{0}}_{b_{0}}\right)
        \end{bmatrix}, \\
        \mbf{A}_{ij} =
        \begin{bmatrix}
            -\mbf{C}_{b_{0}b_{i}}               & \mbf{C}_{b_{0}b_{j}} & -\Delta{t}_{ij}\mbf{1}             \\
            -\mbf{C}_{b_{0}b_{i}}\Delta{t}_{ij} & \mbf{0}              & -\onehalf\Delta{t}_{ij}^{2}\mbf{1}
        \end{bmatrix}.
    \end{gather}
\end{subequations}
By stacking the equations from all consecutive frame pairs, the linear least-squares problem is solved for the initial velocity of each frame and the gravity vector in the IMU frame.
The minimum number of consecutive frames needed to solve for the unknowns is three.
Finally, the angular velocities are computed using finite differences of orientation.

\subsection{Backend Optimization}
\label{subsec:divo-backend}
The states in \eqref{eq:state} are estimated by solving a maximum a posteriori (MAP) problem given a set of measurements in a window as shown in Fig.~\ref{fig:window}.
Assuming the measurements are conditionally independent and corrupted by additive Gaussian noise, the MAP problem is formulated as a nonlinear least-squares problem,
%
\begin{align}
    \hspace{-9pt}
        \mc{X}^{\star} &= \argmin_{\mc{X}} \Vert\mbf{e}^{P}\Vert^{2}_{\mbf{P}_{0}} +
        \sum_{i=k-n}^{k-1} \Big(
        \Vert\mbf{e}^{GP}_{ii+1}\Vert^{2}_{\mbf{Q}_{ii+1}}
        + \Vert\mbf{e}^{I}_{ii+1}\Vert^{2}_{\mbs{\Sigma}_{\mc{I}_{ii+1}}} \nonumber \\
        &+ \sum_{\tau \in \mc{D}_{ii+1}}\rho(\Vert\mbf{e}^{D}_{\tau}\Vert_{\mbs{\Sigma}_{\mc{D}_{\tau}}})
        + \sum_{(j,m) \in \mc{C}_{i}}\rho(\Vert\mbf{e}^{C}_{i,j,m}\Vert_{\mbs{\Sigma}_{\mc{C}_{i}}}) \Big),
\end{align}
where ${\mc{X}}$ is the set of the states in the window defined in~\eqref{eq:state_sliding_window}, ${\Vert\mbf{e}\Vert_{\mbs{\Sigma}}}=\sqrt{\mbf{e}^{\trans}\mbs{\Sigma}\inv\mbf{e}}$ is the Mahalanobis distance, $n$ is the window size, ${\mbf{e}^{P}}$ is the marginalized prior residual with covariance ${\mbf{P}_{0}}$, ${\mbf{e}^{GP}_{ii+1}}$ and ${\mbf{Q}_{ii+1}}$ are the GP prior residual and covariance defined in \eqref{eq:gp_prior_residual} and \eqref{eq:local_covariance}, respectively, and ${\mbf{e}^{I}_{ii+1}}$, ${\mbf{e}^{D}_{\tau}}$, and ${\mbf{e}^{C}_{i,j,m}}$ are the preintegrated IMU, continuous-time DVL, and anchored feature residuals with covariances ${\mbs{\Sigma}_{\mc{I}_{ii+1}}}$, ${\mbs{\Sigma}_{\mc{D}_{\tau}}}$, and ${\mbs{\Sigma}_{\mc{C}_{i}}}$, respectively. 
The set DVL measurement timestamps $\tau$ between $t_{i}$ and $t_{i+1}$ is denoted as ${\mc{D}_{ii+1}}$.
The set of index pairs for feature observations anchored at the $i$-th state is denoted as $\mc{C}_i$, where the $m$-th feature is observed at the $j$-th state.
The Cauchy loss function ${\rho(\cdot)}$~\cite[\S2.5.2]{Barfoot2023State} is used for continuous-time DVL and anchored feature residuals.
\vspace{0.1cm}

\subsubsection{GP WNOA Prior Residual}
\label{subsubsec:divo-backend-prior}
The GP prior residual is derived in Section~\ref{subsec:continuousestimation-gpprior}.
The main difference is that the estimation state \eqref{eq:state} is defined as ${(\mbs{T} \in SE_2(3), \mbs{\omega} \in \mathbb{R}^{3})}$ instead of ${(\mbf{T} \in SE(3), \mbs{\varpi} \in \mathbb{R}^{6})}$.
The body-centric velocity, ${\mbs{\varpi} = \bbm \mbs{\omega}^{\trans} & \mbs{\nu}^{\trans} \ebm^{\trans}}$, can be written as a function of \eqref{eq:state}
\begin{align}
    \mbs{\varpi}(\mbs{T}, \mbs{\omega}) = \begin{bmatrix}
                                              \mbs{\omega}^{\trans} & \left(\mbf{C}_{ab}^{\trans}\mbf{v}^{zw/a}_{a}\right)^{\trans}
                                          \end{bmatrix}^{\trans}.
\end{align}
With this relationship, the mapping from the tangent space of ${(SE_2(3) \times \mathbb{R}^{3})}$ to the tangent space of ${(SE(3) \times \mathbb{R}^{6})}$ is~\cite{Zhang2024GNSSFGO}
\begin{align}
    \label{eq:state_mapping}
    \begin{bmatrix}
        \mbsdel{\xi}^{\mf{se}(3)} \\
        \mbsdel{\varpi}
    \end{bmatrix} \approx
    \begin{bmatrix}
        \f{D\mbf{T}(\mbs{T})}{D\mbs{T}}                    & \mbf{0}                                                 \\
        \f{D\mbs{\varpi}(\mbs{T}, \mbs{\omega})}{D\mbs{T}} & \f{D\mbs{\varpi}(\mbs{T}, \mbs{\omega})}{D\mbs{\omega}}
    \end{bmatrix}
    \begin{bmatrix}
        \mbsdel{\xi}^{\mf{se}_{2}(3)} \\
        \mbsdel{\omega}
    \end{bmatrix},
\end{align}
where ${D\mbf{T}/D\mbs{T}}$ is the Jacobian of the pose matrix with respect to the extended pose, ${D\mbs{\varpi}/D\mbs{T}}$ is the Jacobian of the body-centric velocity with respect to the extended pose, and ${D\mbs{\varpi}/D\mbs{\omega} = \bbm\mbf{1} & \mbf{0}\ebm^{\trans}}$.
Now, the GP prior residual can be linearized with respect to the state defined in \eqref{eq:state}.
\vspace{0.1cm}

\subsubsection{Preintegrated IMU Residual}
\label{subsubsec:divo-backend-imu}
The preintegrated IMU residual for two subsequent states is derived from \eqref{eq:imu_propagation},
\begin{align}
    \mbf{e}^{I}_{ii+1}(\mc{X}_{i}, \mc{X}_{i+1}) & = \mbftilde{U}_{ii+1}^{+} \ominus \mbfcheck{U}_{ii+1}(\mbs{T}_{i}, \mbs{T}_{ii+1}), \\
    \mbftilde{U}_{ii+1}^{+} &= \mbftilde{U}_{ii+1}\oplus\f{D\mbf{U}_{ii+1}}{D\mbf{b}_{i}}\delta\mbf{b}_{i},
\end{align}
where ${\mbftilde{U}_{ii+1}}$ is the preintegrated IMU measurement from \eqref{eq:imu_propagation}, ${{D\mbf{U}_{ii+1}}/{D\mbf{b}_{i}}}$ is the Jacobian of the preintegrated IMU measurement with respect to the IMU biases, ${\delta\mbf{b}_{i} = \mbf{b}_{i} - \mbfbar{b}_{i}}$ is the difference between the current bias and the bias used for preintegration, and ${\mbfcheck{U}_{ii+1}(\mbs{T}_{i}, \mbs{T}_{i+1}) = (\mbf{G}_{ii+1}\mbs{T}_{i})\inv\mbs{T}_{i+1}}$.
\vspace{0.1cm}

\subsubsection{Continuous-time DVL Residual}
\label{subsubsec:divo-backend-dvl}
The DVL residual is formulated by comparing the DVL measurement to the predicted DVL velocity at the measurement time ${\tau}$,
\begin{align}
    \mbf{e}^{D}_{\tau}\left(\mc{X}_{\tau}\right) & = \mbftilde{v}^{d_{\tau}a/a}_{d_{\tau}} - \mbf{g}_{d}(\mc{X}_{\tau}),
\end{align}
where ${\mbftilde{v}^{d_{\tau}a/a}_{d_{\tau}}}$ is from \eqref{eq:dvl_velocity}, and ${\mbf{g}_{d}(\cdot)}$ is defined as
\begin{equation}
    \begin{split}
        \mbf{g}_{d}(\mc{X}_{\tau}) = \mbf{C}_{bd}^{\trans}\left(\mbf{C}_{ab_{\tau}}^{\trans}\left(\mbf{v}^{b_{\tau}a/i}_{a} - \left(\mbs{\omega}^{ai}_{a}\right)^{\times}\mbf{r}^{b_{\tau}a}_{a}\right)\right. \\
        \left.+ \left(\mbs{\omega}^{b_{\tau}i}_{b_{\tau}} - \mbf{C}_{ab_{\tau}}^{\trans}\mbs{\omega}^{ai}_{a}\right)^{\times}\mbf{r}^{db}_{b}\right) + \mbs{\eta}^{\textrm{dvl}}_{d_{\tau}},
    \end{split}
\end{equation}
with ${(\mbf{C}_{bd}, \mbf{r}^{db}_{b})}$ as the extrinsic calibration between the DVL and IMU and ${\mbs{\eta}^{\textrm{dvl}}_{d_{\tau}} \sim \mc{N}(\mbf{0}, \mbs{\Sigma}^{\textrm{dvl}}_{\tau})}$ as the measurement noise.
Without Earth's rotation effect, the model is simplified to
\begin{subequations}
    \begin{align}
        \mbf{g}_{d}(\mc{X}_{\tau}) & = \mbf{C}_{bd}^{\trans}\left(\mbf{C}_{ab_{\tau}}^{\trans}\mbf{v}^{b_{\tau}a/a}_{a} + \left(\mbs{\omega}^{b_{\tau}a}_{b_{\tau}}\right)^{\times}\mbf{r}^{db}_{b}\right) + \mbs{\eta}^{\textrm{dvl}}_{d} \\
                                    & = \mbf{C}_{bd}^{\trans}\begin{bmatrix}
                                                                 -\left(\mbf{r}^{db}_{b}\right)^{\times} & \mbf{1}
                                                             \end{bmatrix}\mbs{\varpi}(\tau) + \mbs{\eta}^{\textrm{dvl}}_{d},
    \end{align}
\end{subequations}
where the body-centric velocity at time ${\tau}$ is obtained using the GP interpolation as described in Section~\ref{subsec:continuousestimation-gpinterpolation}.
Then, the DVL residual is linearized with respect to the state at time ${\tau}$ using the chain rule with \eqref{eq:state_mapping}.
This allows the DVL measurements to be incorporated at any time between two camera frames, yet maintain the number of columns of the Jacobian to be the same, yielding a computationally efficient optimization.
\vspace{0.1cm}

\subsubsection{Anchored feature Residual}
\label{subsubsec:divo-backend-visual}
The anchored feature residual is formulated as
\begin{align}
    \mbf{e}^{C}_{i,j,m}(\mc{X}_{i}, \mc{X}_{j}) & = \begin{bmatrix}
                                                          u_{m} & v_{m}
                                                      \end{bmatrix}^{\trans}_{j} - \mbs{\pi}\left(\mbf{T}_{ab_{j}}^{-1}\mbf{T}_{ab_{i}}\mbftilde{r}^{\ell_{m} b_{i}}_{b_{i}}\right),
\end{align}
where ${\bbm u_{m} & v_{m} \ebm^{\trans}_{j}}$ is the observed pixel coordinate of the ${m}$-th feature in the ${j}$-th image, ${\mbftilde{r}^{\ell_{m}b_{i}}_{b_{i}}}$ is the 3D position of the feature anchored at the ${i}$-th image in homogeneous coordinates, and ${\mbs{\pi}(\cdot)}$ is the pinhole camera projection function.
Once the features are marginalized, the same feature that is tracked in the newest image is re-initialized at the current state by triangulation.

\subsection{Implementation Details}
\label{subsec:divo-implementation}
In this work, the window size is set to \SI{3}{\second} in the experiments.
When the window is full, the oldest state is marginalized out~\cite{Sibley2010Sliding}.
All the sensor extrinsics, intrinsics, and time offsets are pre-calibrated and kept constant during the experiments.
Recent work have explored the idea of calibrating the WNOA hyperparameter ${\mbc{Q}}$ using supervised-learning approaches.
However, high variation vehicle dynamics during operation makes it difficult to collect representative training data.
Thus, the power spectral density tuning is left for future work, and a constant value is used throughout the experiments.
The system is implemented in \texttt{C++} with \texttt{ROS2}~\cite{ROS2}.
The graph optimization is solved using the incremental smoothing algorithm iSAM2~\cite{Kaess2011iSAM2} in \texttt{GTSAM}~\cite{gtsam}.
The experiments are run on a laptop with an Intel i9-13900H, $20$~cores and \SI{16}{\giga\byte} of RAM.
All GPU-based operations in testing were computed on a NVIDIA RTX 2000 Ada Generation Laptop GPU with CUDA 13.0, CuDNN 9.0, and ONNX Runtime 1.24.1.
The RTX 2000 has \SI{8}{\giga\byte} of dedicated GPU memory.
\begin{figure}[t]
    \centering
    \begin{subfigure}[t]{\linewidth}
        \centering
        \includegraphics[width=0.9\linewidth]{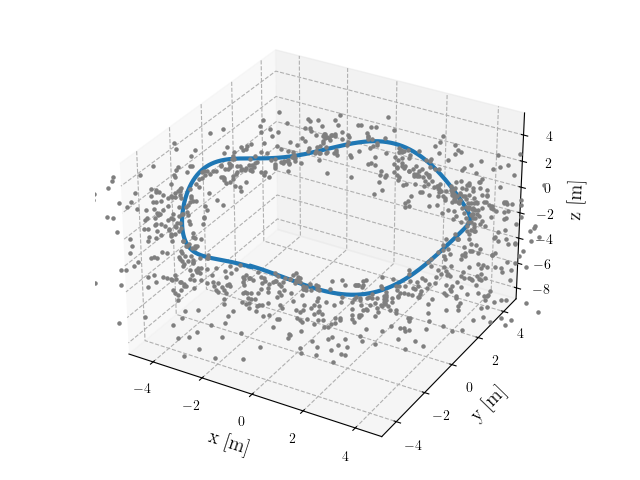}
        \caption{Circular trajectory with sinusoidal vertical motion.}
        \label{fig:sim}
    \end{subfigure}
    \begin{subfigure}[t]{\linewidth}
        \centering
        \includegraphics[width=0.9\linewidth]{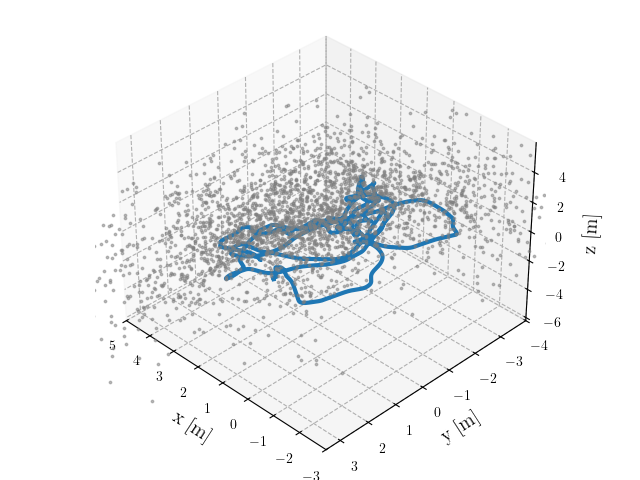}
        \caption{Vicon room sequence from the EuRoC dataset.}
        \label{fig:euroc}
    \end{subfigure}
    \caption{
        Simulation setup with two different trajectories.
        The vehicle is equipped with a forward-facing camera, IMU, and DVL.
        The camera observes landmarks (grey dots) in the environment.
    }
    \label{fig:sim_setup}
\end{figure}

In all evaluations on real data, the proposed system runs with a matching window length of ${n=10}$ frames with a static \SI{2}{px} noise for all visual features, disparity threshold ${d_n = \SI{10}{px}}$, and minimum matched features ${m_n = \SI{10}{}}$. This implementation runs at \SI{8}{fps} on the evaluation setup, though this varies with the feature extraction density. The proposed system runs in real-time for all presented data.

\section{Simulations \& Real-world Experiments}
\label{sec:results}
The performance of the proposed system is evaluated in both simulation and real-world experiments for visual-inertial and acoustic-visual-inertial sensor suites.

\subsection{Simulation Experiments}
\label{subsec:results-simulation}

\subsubsection{Setup}
\label{subsubsec:results-simulation-setup}
Two simulation settings are considered here: (1) a vehicle following ${5}$ revolutions of \SI{5}{\meter} radius circular trajectory with a sinusoidal vertical motion and a vehicle speed of \SI{1}{\meter\per\second}, and (2) a vehicle following a trajectory of the Vicon room sequence from the EuRoC dataset~\cite{Burri2016EuRoC}.
The EuRoC dataset is chosen here because it provides a challenging trajectory with rapid motions and varying speeds compared to typical underwater vehicle motions.
The vehicle is assumed to be equipped with a DVL, stereo camera, and IMU.
The simulation setup is depicted in Fig.~\ref{fig:sim_setup} with trajectories shown in blue lines and landmarks in grey dots.
\begin{table}[t]
    \centering
    \begin{tabular}{c|c|c|c|c}
        \toprule
               & \multicolumn{4}{c}{Circular Trajectory} \\
        \hline
        System & ATE (\SI{}{\meter}) & RMSE (\SI{}{\meter}) & ATE (\SI{}{\degree}) & RMSE (\SI{}{\degree}) \\
        \hline
        DIO    & 0.2369          & 0.2339          & 0.0262          & 0.0285          \\
        VIO    & 0.1161          & 0.1162          & \textbf{0.0129} & \textbf{0.0139} \\
        DIVO   & \textbf{0.1148} & \textbf{0.1132} & 0.0136          & 0.0146          \\
        \midrule
               & \multicolumn{4}{c}{EuRoC Vicon Room} \\
        \hline
        System & ATE (\SI{}{\meter}) & RMSE (\SI{}{\meter}) & ATE (\SI{}{\degree}) & RMSE (\SI{}{\degree}) \\
        \hline
        DIO    & 0.1282          & 0.1247          & 0.0316          & 0.0340          \\
        VIO    & 0.0738          & 0.0686          & 0.0165          & 0.0170          \\
        DIVO   & \textbf{0.0717} & \textbf{0.0673} & \textbf{0.0163} & \textbf{0.0169} \\
        \bottomrule
    \end{tabular}
    \caption{
        The mean position and rotation ATE and RMSE over 50 Monte Carlo runs for both trajectories are shown.
        The proposed DIVO system outperforms both DIO and VIO in terms of estimation accuracy.
    }
    \label{tab:sim_results}
\end{table}
\begin{figure}[t]
    \centering
    \includegraphics[width=0.95\linewidth]{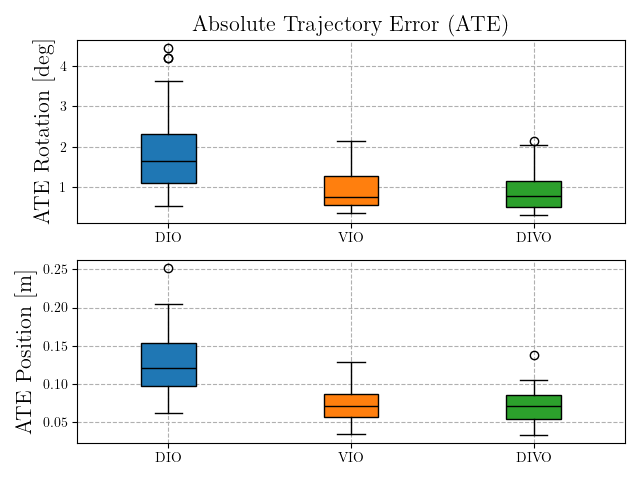}
    \caption{
        Boxplot of ATEs over 50 Monte Carlo runs for EuRoC Vicon Room.
        The proposed DIVO system outperforms both DIO and VIO in terms of rotation and position estimation accuracy.
    }
    \label{fig:sim_boxplot}
\end{figure}
The simulated camera measurements are sampled at \SI{10}{\hertz}.
The number of landmark observations per frame is limited to 20 to simulate a low-feature underwater environment with maximum depth of \SI{5}{\meter}.
To simulate a realistic feature-tracker, the features are corrupted with isotropic Gaussian noise. 
The simulated IMU measurements are computed at \SI{100}{\hertz} using the cubic B-spline~\cite{Geneva2020OpenVINS} and corrupted by white noise and random walk biases.
To simulate the asynchronous DVL measurements, the velocities are sampled and corrupted by white noise at \SI{6}{\hertz} with a \SI{0.033}{\second} time offset.

Three systems are tested for consistency and accuracy: (1) DIO using DVL and IMU, optimized at every DVL timestamp, (2) VIO using camera, IMU, and some DVL measurements that are time-synchronized with the camera images, and (3) DIVO using camera, IMU, and DVL measurements with the proposed continuous-time DVL model.
The hyperparameter ${\mbc{Q}}$ for the WNOA GP prior is set to ${\diag{(\mbc{Q})} = \{100, 100, 100, 10, 10, 10\}}$.
A Monte Carlo analysis with 50 runs is performed.
\vspace{0.1cm}

\subsubsection{Results}
\label{subsubsec:results-simulation-results}
The performance is evaluated in terms of estimation accuracy, consistency, and computational efficiency.
Estimation accuracy is evaluated using the absolute trajectory error (ATE) and root-mean-square error (RMSE) metric as presented in~\cite{Sturm2012Benchmark}.
A summary of the mean ATE and RMSE over 50 Monte Carlo runs is presented in Table~\ref{tab:sim_results}.
As expected, the proposed system outperforms both DIO and VIO in terms of estimation accuracy for both trajectories.
The VIO system performs comparably with DIVO because it also uses DVL data that is measured at the same time as the cameras, which is rare in practice due to the asynchronous nature of the sensors.
\begin{figure}[t!]
    \centering
    \includegraphics[width=0.9\linewidth]{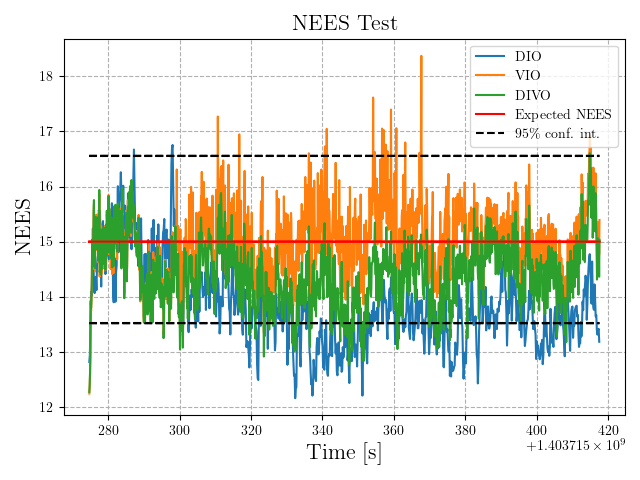}
    \caption{
        NEES averaged over 50 Monte Carlo runs for the EuRoC Vicon Room.
        The results show that all systems are consistent, including the proposed DIVO.
    }
    \label{fig:sim_nees}
\end{figure}
\begin{figure}[t]
    \centering
    \includegraphics[width=0.95\linewidth]{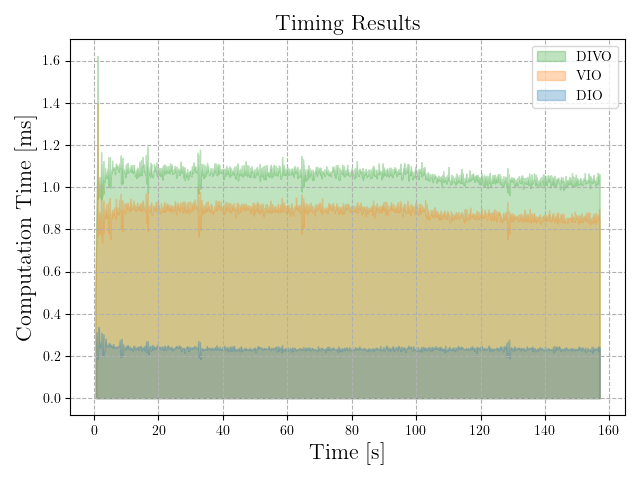}
    \caption{
        Computation time per optimization.
        DIVO has a comparable computation time to VIO.
    }
    \label{fig:sim_timing}
\end{figure}

A distribution of the ATEs for all Monte Carlo runs is shown in Fig.~\ref{fig:sim_boxplot}.
Despite VIO outperforming in mean rotational ATE, DIVO has a lower variance in both position and rotation ATEs.
This indicates that the proposed DIVO system is more robust to different noise realizations.

The optimal estimate from a consistent estimator should not only provide the accurate result, but also correctly indicate the quality of the estimate.
The average normalized estimation error squared (NEES)~\cite{Bar-Shalom2002Estimation} is computed for the extended pose and biases to evaluate the estimator consistency.
Figure~\ref{fig:sim_nees} shows the NEES results for the EuRoC Vicon Room.
The results show that all systems are consistent including the proposed DIVO.

The computational efficiency is evaluated by measuring the computation time per optimization.
The timing results at every optimization step are shown in Fig.~\ref{fig:sim_timing}.
The results show that DIVO has a comparable computation time to VIO despite incorporating the motion prior and continuous-time DVL model with GP interpolation.

\subsection{State-of-the-art Benchmarking}
\label{subsec:comp-algos}

To define baseline results for real-world experiments, a collection of state-of-the-art algorithms for visual-inertial and acoustic-visual-inertial modalities are evaluated.
Each algorithm's configuration source, if any, is specified in its overview, and is the default open-source configuration if unspecified.
Absolute and relative trajectory errors are reported on the overlapping regions of all methods for a fair comparison. 

The following visual-inertial algorithms are used to compare against the proposed visual-inertial solution.
\begin{enumerate}
    \item \textbf{ORB-SLAM3}~\cite{camposORBSLAM3AccurateOpenSource2021} presents a keyframe-based graph optimization framework with a multi-mapping module Atlas and DBoW2~\cite{galvez-lopezBagsBinaryWords2012} based place recognition for loop closure and map merging.
    Results are generated with a relaxed FAST corner response threshold to accommodate poor visibility in a subsea environment.
    Tracking parameters were inherited from AQUA-SLAM to achieve better underwater performance.
    \item \textbf{OKVIS2-X}~\cite{bocheOKVIS2XOpenKeyframebased2025} presents a keyframe-based fixed-lag smoother with DBoW2~\cite{galvez-lopezBagsBinaryWords2012} based place recognition for loop closure and Unimatch-based volumetric submapping~\cite{xuUnifyingFlowStereo2023}.
    Results are presented with and without submapping, with loop closure enabled in both cases.
    Modifications were made to the frontend parameters, place similarity threshold, and loop closure error heuristics to allow the feature tracking and place recognition module to function in a subsea environment.
    The occupancy submapping was run at 5 $\si{cm}$ voxel size. OKVIS2-X is made single-threaded for all sequences to avoid inter-process communication introducing randomness to the evaluation.
\end{enumerate}
\begin{figure}[t]
    \centering
    \begin{subfigure}[t]{0.49\linewidth}
        \centering
        \includegraphics[width=\linewidth]{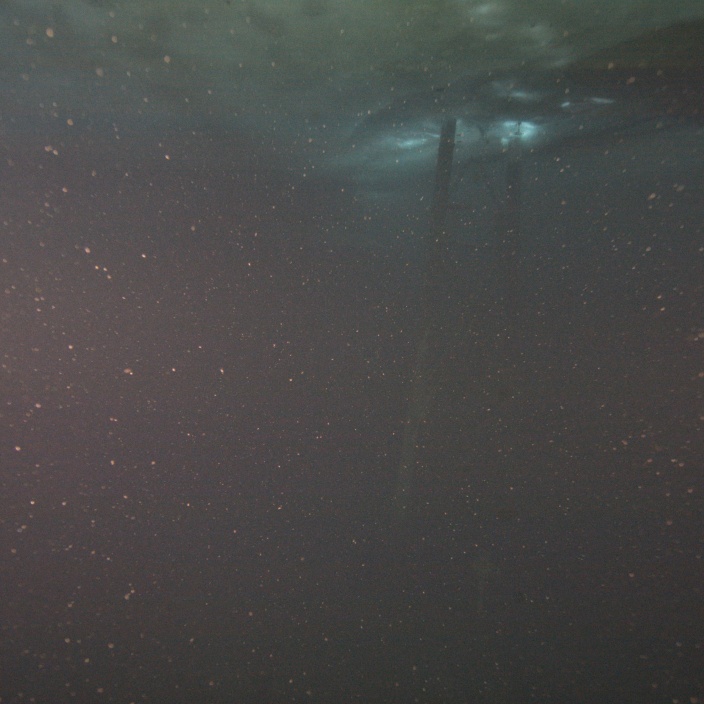}
        \caption{Far from target}
    \end{subfigure}
    \begin{subfigure}[t]{0.49\linewidth}
        \centering
        \includegraphics[width=\linewidth]{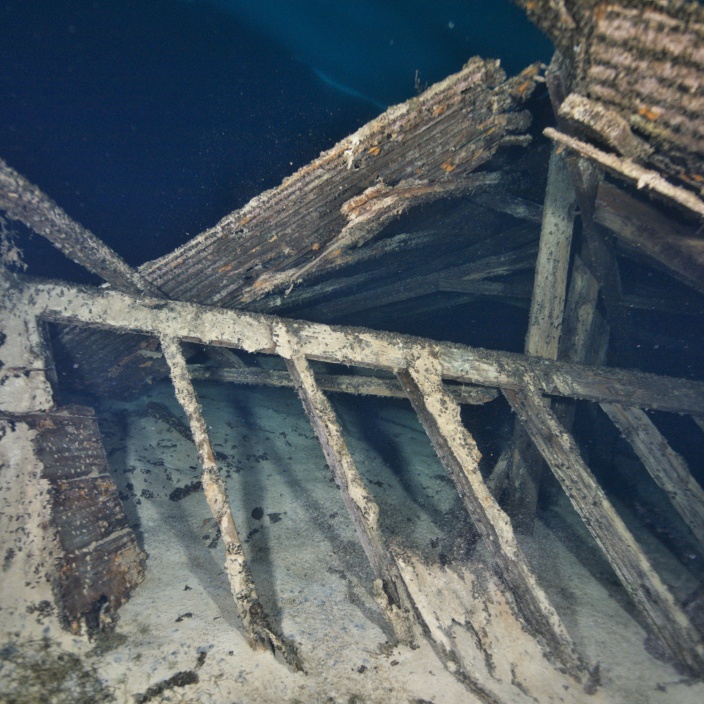}
        \caption{Close to target}
    \end{subfigure}
    \vskip\baselineskip
    \vspace{-0.2cm}
    \begin{subfigure}[b]{0.49\linewidth}
        \centering
        \includegraphics[width=\linewidth]{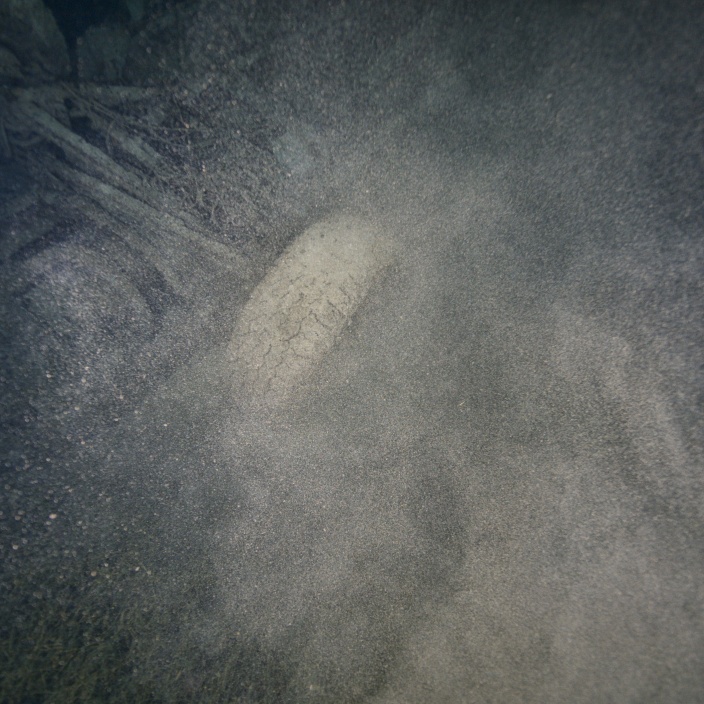}
        \caption{Dynamic Particles}
    \end{subfigure}
    \begin{subfigure}[b]{0.49\linewidth}
        \centering
        \includegraphics[width=\linewidth]{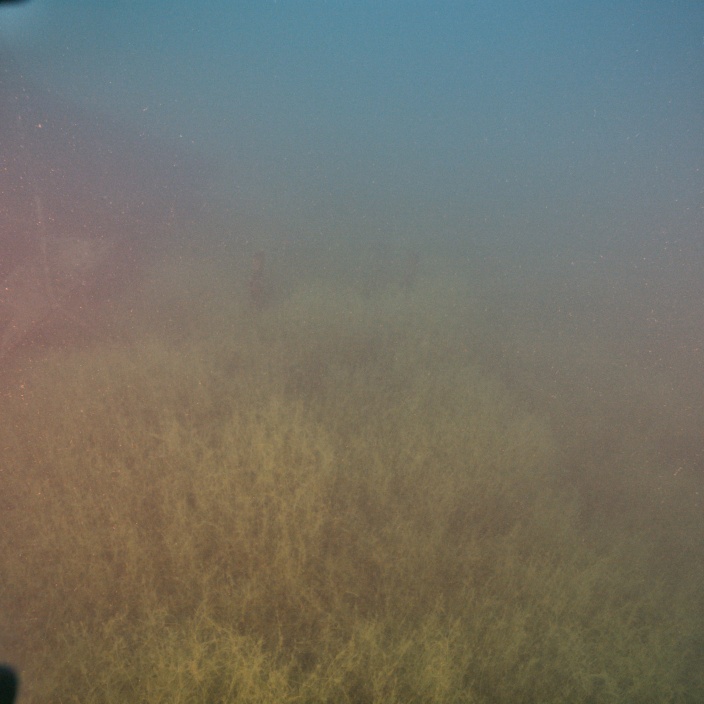}
        \caption{Unstructured Environments}
    \end{subfigure}
    \vskip\baselineskip
    \vspace{-0.2cm}
    \begin{subfigure}[b]{0.49\linewidth}
        \centering
        \includegraphics[width=\linewidth]{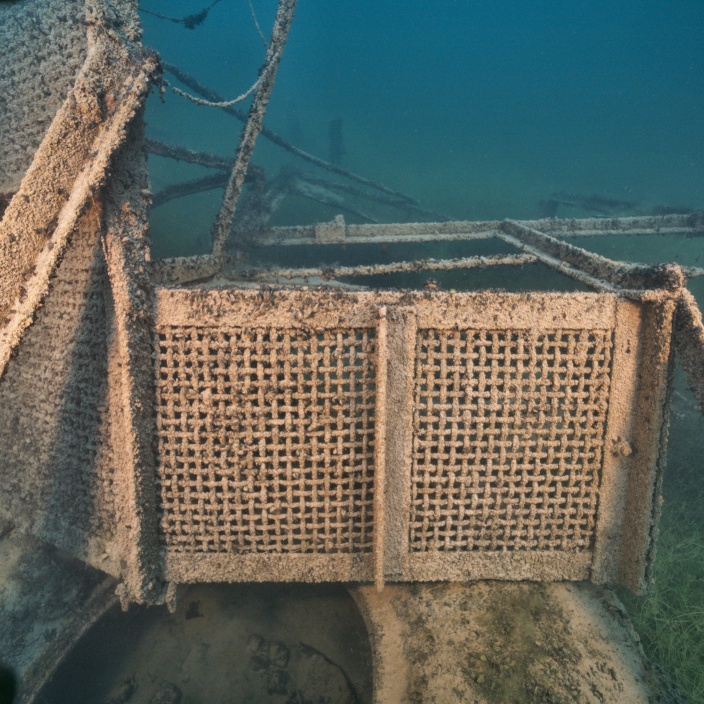}
        \caption{Auto white balance on}
    \end{subfigure}
    \begin{subfigure}[b]{0.49\linewidth}
        \centering
        \includegraphics[width=\linewidth]{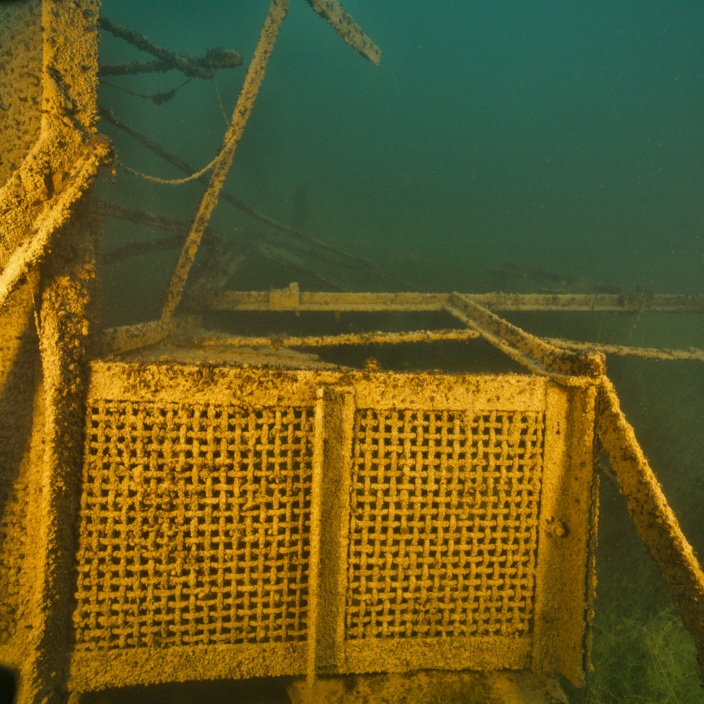}
        \caption{Auto white balance off}
    \end{subfigure}
    \caption{
        Examples showing different imaging conditions.
        Top left: Far from the target.
        Top right: Close to the target.
        Middle left: Dynamic sand particles.
        Middle right: Visually degenerate environment.
        Bottom left: Image with auto white balance turned on.
        Bottom right: Image with auto white balance turned off.
    }
    \label{fig:image_setup}
\end{figure}
The following acoustic-aided algorithms are used to compare against the proposed acoustic-visual-inertial solution. 
\begin{enumerate}
    \item \textbf{MSCKF-DVIO}~\cite{Zhao2023Tightly} presents a multi-modal sensor fusion based on the MSCKF framework~\cite{Mourikis2007MSCKF} to address the challenging underwater image conditions by fusing DVL, IMU, monocular camera, and pressure sensor.
    Despite being a monocular solution, this work uses the DVL to constrain the camera feature depth using the range measurements from its beams on top of it being the linear velocity sensor.
    Due to the lack of open-source state-of-the-art solution that incorporates a stereo, IMU, and a DVL, this method is chosen to be compared against.
    The system's navigation states were initialized with the estimated output of the proposed DIVO system.
    The frontend parameters, namely, the number of points per image, and the minimum distance per feature is modified to match with the proposed method.
    \item \textbf{AQUA-SLAM}~\cite{Xu2025AQUA-SLAM} fuses the ORB-SLAM3~\cite{camposORBSLAM3AccurateOpenSource2021} visual-inertial framework with the DVL-IMU preintegration~\cite{Vial2025Lie} to yield a tightly-coupled acoustic-visual-inertial SLAM algorithm.
    In addition, AQUA-SLAM can perform online sensor calibration for both DVL-camera-IMU extrinsics and DVL transducer misalignment in real time.
    Tracking and DVL parameters were inherited from those used for the TANK dataset~\cite{Xu2025Tank}.
    The current version of the open-source code fails when running with long sequences, thus a data loading modification has been made, which will be open-sourced.
\end{enumerate}
The proposed system and MSCKF-DVIO are causal, meaning they only take measurements up to the current time. All other algorithms are non-causal and present a final loop-closed trajectory. Further, all algorithms are evaluated with no real-time constraints to avoid data drops. All algorithm configurations will be open-sourced to encourage transparency.

\subsection{Real-world Experiments}
\label{subsec:results-experimental}
\begin{table}[t]
    \caption{
        Quarry dataset sequences for quantitative analysis.
    }
    \label{tab:quarry}
    \centering
    \begin{tabular}{c c c c c}
        \toprule
        Sequence
        & Duration
        & Distance
        & Max speed
        & Floaters \\
        \midrule
        Conveyor1 & \SI{20}{\minute}\SI{40}{\second} & \SI{148.40}{\meter} & \SI{1.40}{\meter\per\second} & Low\\
        Conveyor2 & \SI{25}{\minute}\SI{59}{\second} & \SI{222.62}{\meter} & \SI{1.82}{\meter\per\second} & Low\\
        Truck1    & \SI{23}{\minute}\SI{39}{\second} & \SI{185.51}{\meter} & \SI{0.66}{\meter\per\second} & Low\\
        Truck2    & \SI{9}{\minute}\SI{43}{\second}  & \SI{89.05}{\meter} & \SI{0.93}{\meter\per\second} & High\\
        \bottomrule
    \end{tabular}
\end{table}

\subsubsection{Setup}
\label{subsubsec:results-experimental-setup}
The Quarry dataset is collected in a freshwater with a DeepTrekker Revolution ROV equipped with Voyis Discovery stereo camera running at \SI{3}{\hertz}, Microstrain 3DM-GX5-25 IMU at \SI{100}{\hertz}, and Waterlinked A50 DVL at \SI{10}{\hertz} as shown in Fig.~\ref{fig:frames}.
The image and IMU data were transferred using a fiber optic cable, and other sensors and vehicle information were transferred using another tether under User Datagram Protocol (UDP).
Despite the DVL operating at \SI{10}{\hertz}, the UDP caused some data drops resulting in the operating frequency between \SI{6} to \SI{8}{\hertz}.
All sensor clocks were NTP synchronized.

Ground truth poses are provided by the photogrammetry software, Agisoft Metashape. To the best of the authors' knowledge, there is no open-source subsea dataset with this sensor suite and photogrammetric ground truth. The data used herein will be released in a future publication as a standalone dataset paper.

Four sequences are collected for the quantitative analysis.
The sequences are obtained under varying scenarios, including different asset types and different speeds.
A brief summary of the data sequences is shown in Table~\ref{tab:quarry}.
Proximity conditions are shown in Fig.~\ref{fig:image_setup}.
When the ROV is close to a target, it captures images with good visibility of the scene that may result in more visual features for tracking.
Conversely, when the ROV is far from the target, the visibility is poor, making it challenging to extract and track visual features.

In some of the sequences, there are regions of the trajectory where sand dust from the Quarry floor is propelled up by the thrusters of the ROV as shown in Fig.~\ref{fig:image_setup}.
The presence of floating particulate matter in the water column makes the algorithm more exposed to outlier tracking matching.

White balance is the process of adjusting colors to ensure the images appear true-to-life under different lighting conditions by preventing unnatural color casts as shown in Fig.~\ref{fig:image_setup}.
It compensates for the color temperature of the light source, which can vary underwater due to factors such as depth and water clarity.
All sequences have the automatic white balance turned on.
Two different sensor combinations are compared here: VIO using camera and IMU and DIVO using camera, IMU, and DVL with the proposed continuous-time DVL model.
\vspace{0.1cm}
\subsubsection{Sensor Calibration}
The camera calibration is two-fold.
First, a proprietary camera calibration software, is run to provide a monocular and stereo intrinsics calibration. This calibrated camera parameters are used to undistort and rectify images at the image acquisition, which results in a rectified stereo pair. Then, Kalibr~\cite{rehderExtendingKalibrCalibrating2016} computes the camera-IMU spatiotemporal extrinsics using the rectified intrinsics and the stereo baseline. Finally, the calibrated camera intrinsics and extrinsics are used in all visual-inertial and acoustic-visual-inertial algorithms.

The DVL-IMU extrinsics are calculated from the CAD drawings of the sensor rig.
\vspace{0.1cm}
\begin{figure*}[t]
    \centering
    \includegraphics[width=0.8\linewidth]{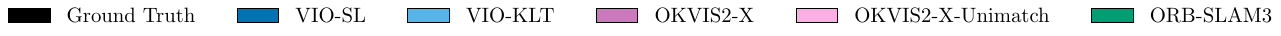}
    \begin{subfigure}[t]{0.5\linewidth}
        \centering
        \includegraphics[width=\linewidth]{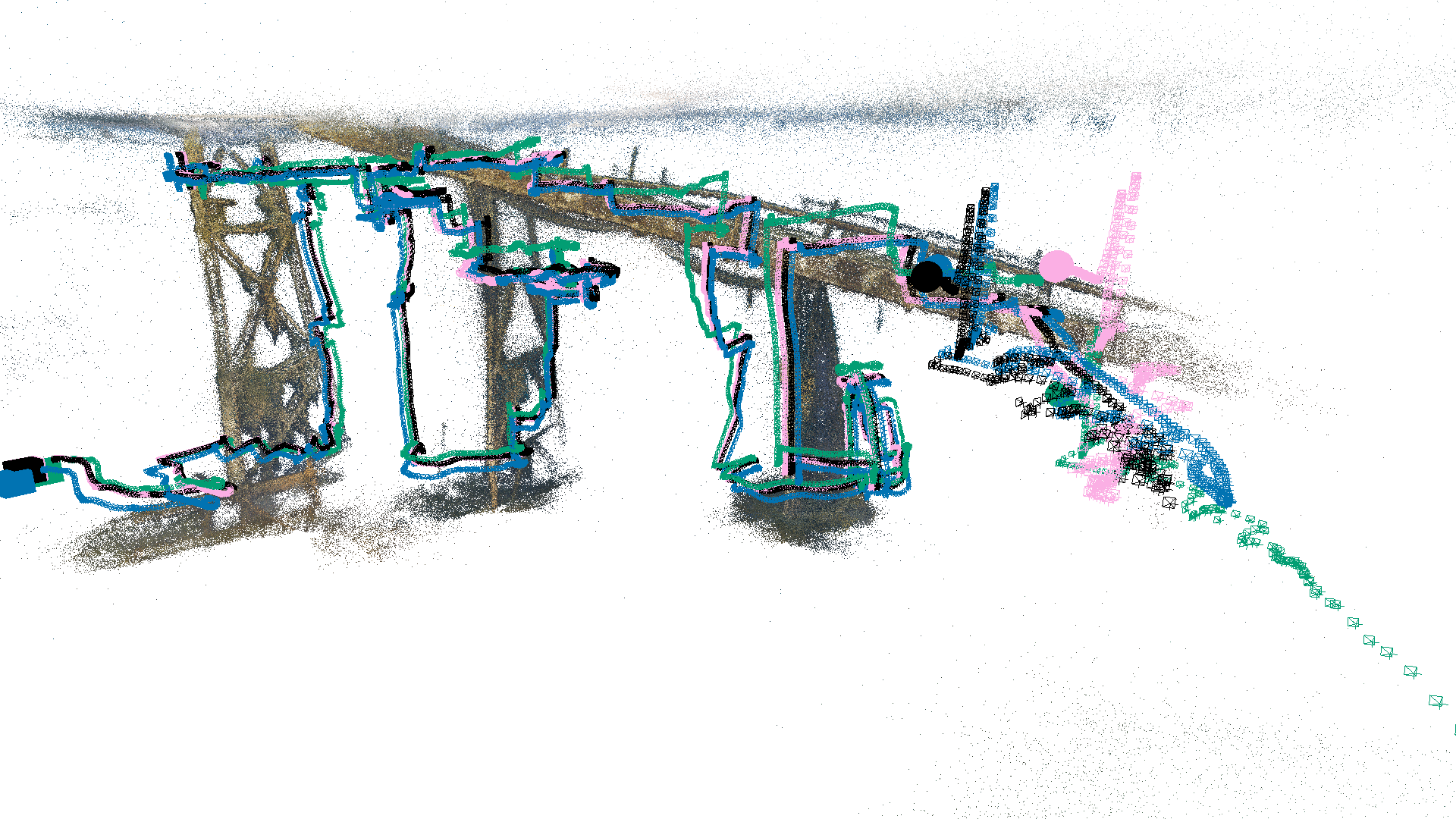}
        \caption{VIO-SL vs. OKVIS2-X-Unimatch vs. ORB-SLAM3 on Conveyor1.}
        \label{fig:vio_traj_conv1}
    \end{subfigure}%
    \begin{subfigure}[t]{0.5\linewidth}
        \centering
        \includegraphics[width=\linewidth]{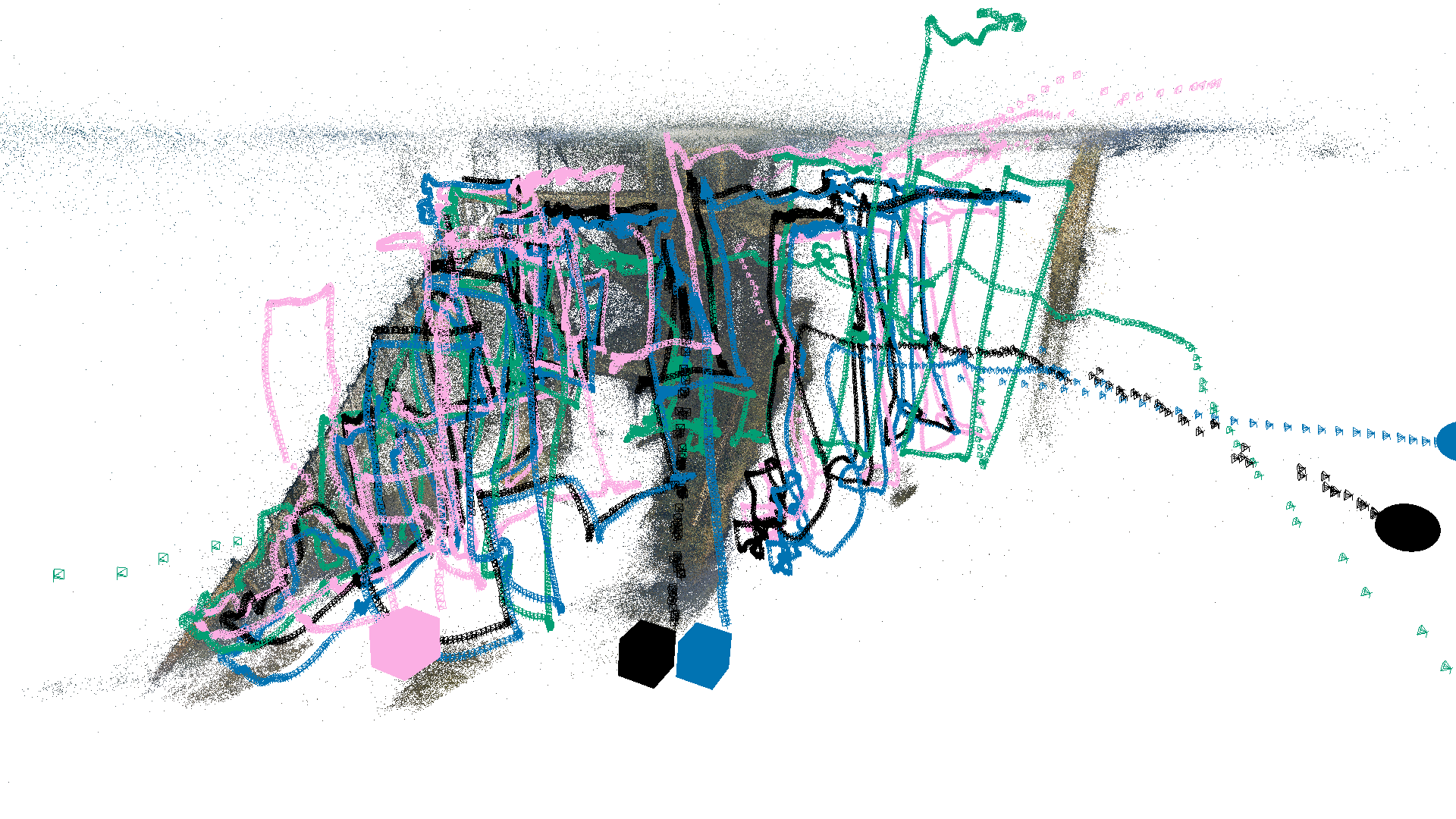}
        \caption{VIO-SL vs. OKVIS2-X-Unimatch vs. ORB-SLAM3 on Conveyor2.}
        \label{fig:vio_traj_conv2}
    \end{subfigure}
    \vskip\baselineskip
    \begin{subfigure}[t]{0.5\linewidth}
        \centering
        \includegraphics[width=\linewidth]{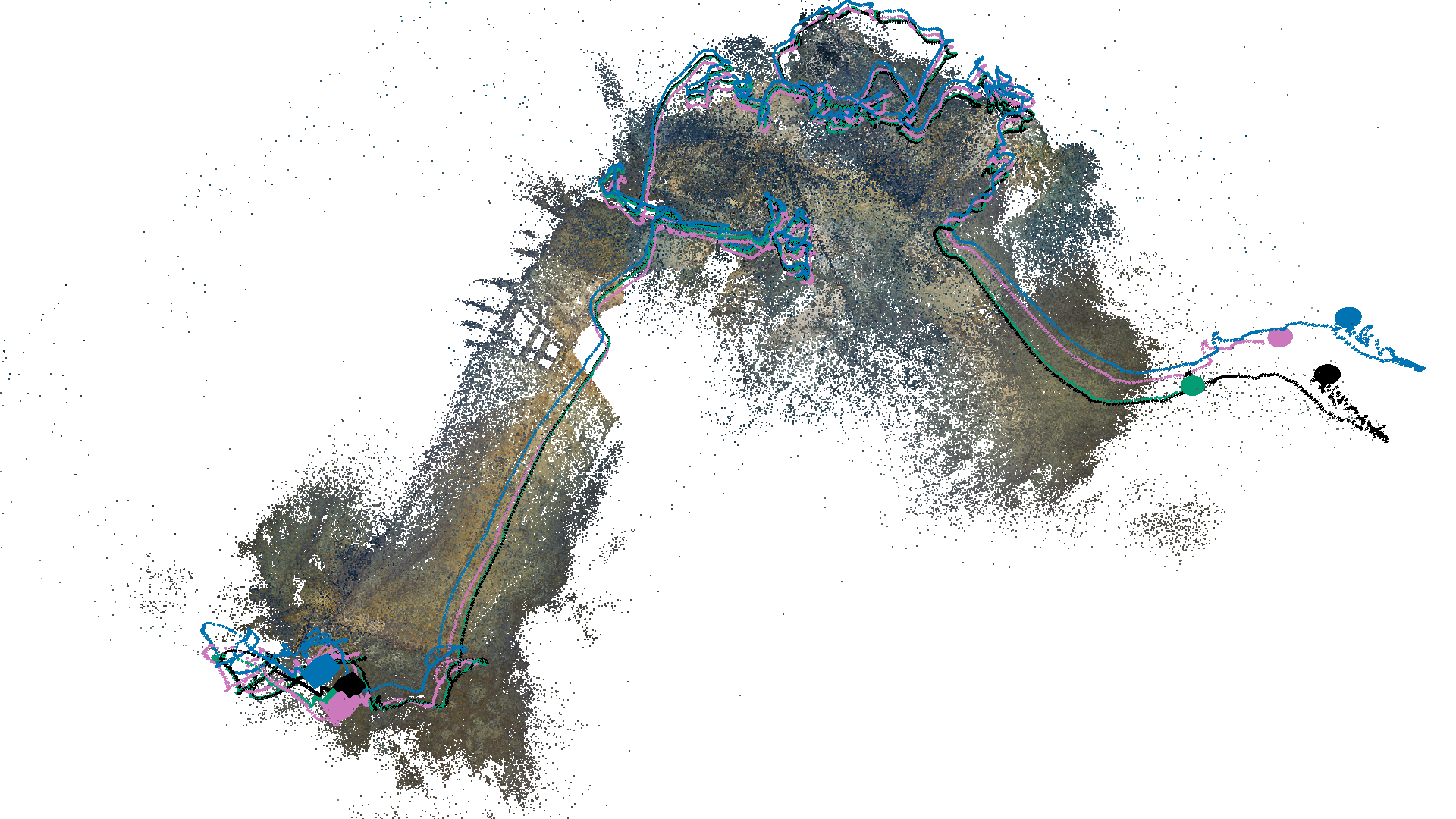}
        \caption{VIO-SL vs. OKVIS2-X vs. ORB-SLAM3 on Truck1.}
        \label{fig:vio_traj_truck1}
    \end{subfigure}%
    \begin{subfigure}[t]{0.5\linewidth}
        \centering
        \includegraphics[width=\linewidth]{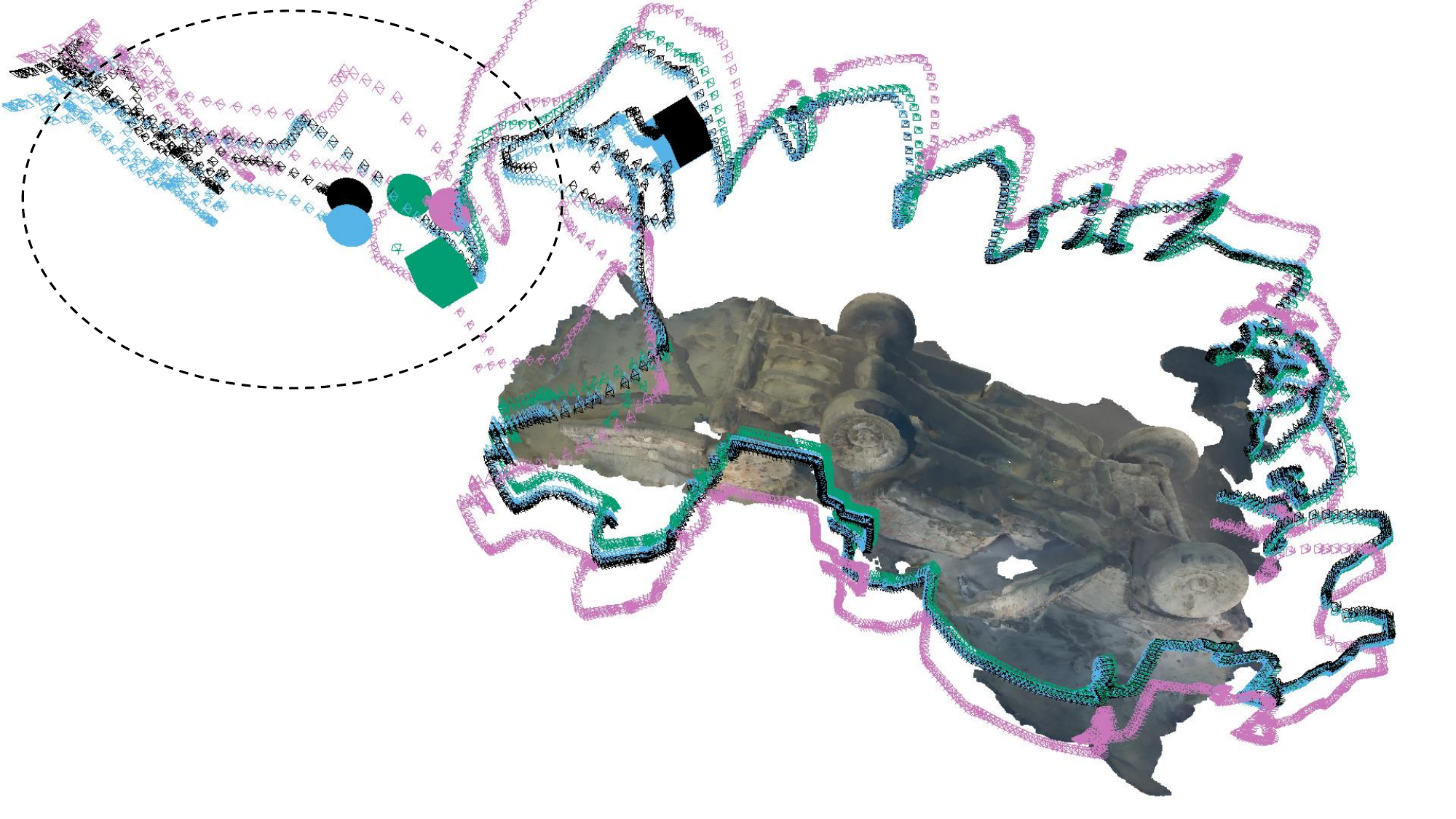}
        \caption{VIO-KLT vs. OKVIS2-X vs. ORB-SLAM3 on Truck2.}
        \label{fig:vio_traj_truck2}
    \end{subfigure}
    \vskip\baselineskip
    \begin{subfigure}[t]{\linewidth}
        \centering
        \includegraphics[width=.3\linewidth]{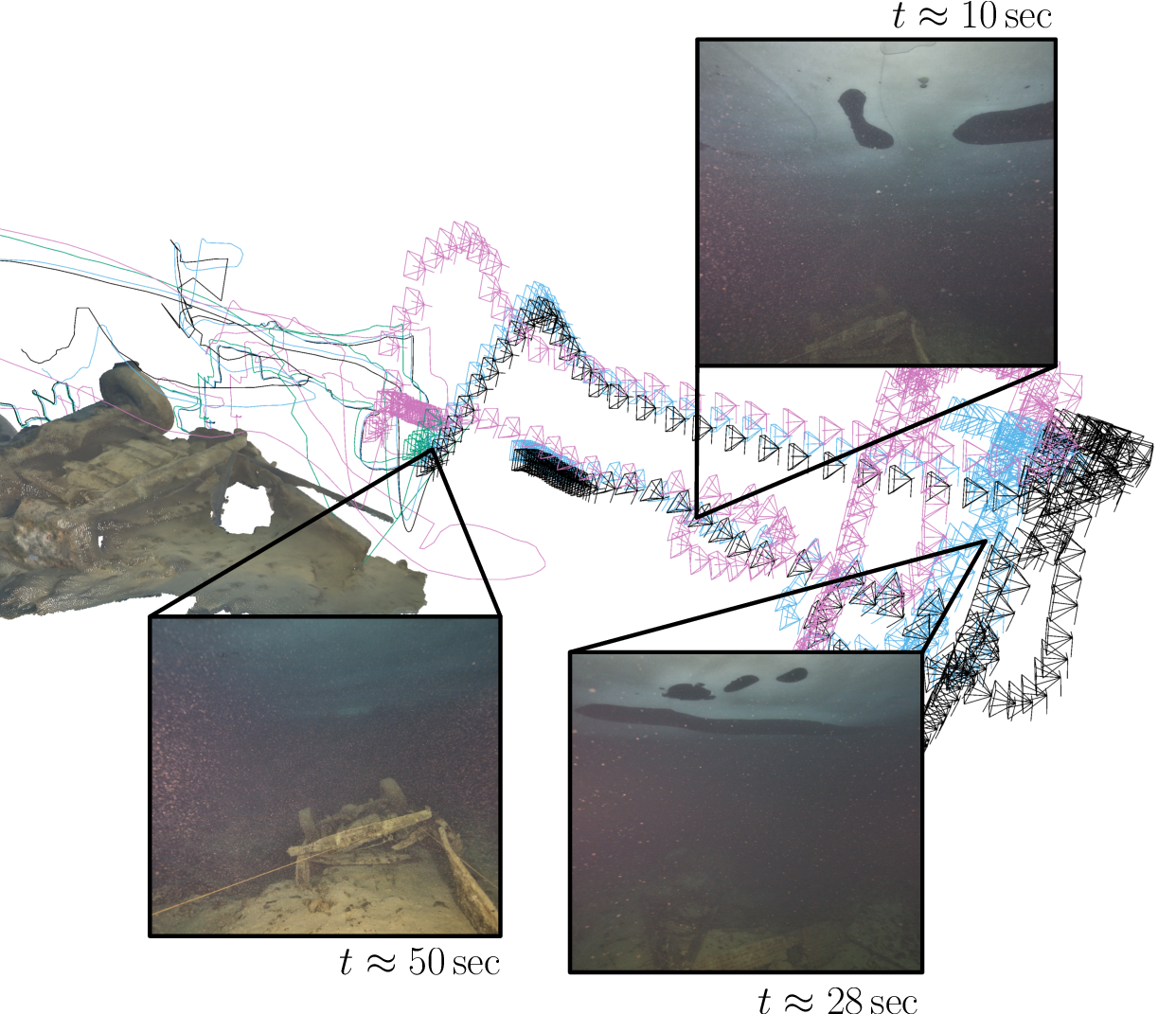}
        \caption{VIO-KLT (\textcolor{cyan}{blue}), OKVIS2-X (\textcolor{violet}{purple}), and ORB-SLAM3 (\textcolor{teal}{green}) in the initial region of Truck2, circled in (d). ORB-SLAM3 does not track for the first $\approx \SI{54}{\sec}$.
        }
        \label{fig:vio_traj_truck2_init}
    \end{subfigure}
    \caption{
        The trajectory of visual-inertial algorithms overlaid to the ground truth generated by Agisoft Metashape (black). Spheres denote starting positions and boxes denote ending positions for each method.
    }
    \label{fig:vio_traj}
\end{figure*}

\subsubsection{Ground Truth Generation Pipeline}
Agisoft Metashape is used to generate metric photogrammetry and camera poses from the undistorted and rectified stereo image pairs. First, an initial sparse feature set and camera trajectory is generated from a proprietary structure-from-motion (SfM) algorithm. The feature set is then culled by thresholding on point uncertainty and minimum correspondences. Lastly, outlying points are manually filtered out of the cloud by human inspection. The camera trajectory is re-optimized at each refinement step, and this process is repeated for each sequence of the Quarry dataset. Post-refinement, the IMU trajectory is retrieved using the computed camera-IMU extrinsics. A visualization of the metric reconstructions and camera trajectories are shown in Fig.~\ref{fig:vio_traj} and Fig.~\ref{fig:divo_traj}.
\renewcommand{\arraystretch}{1.2}
\begin{table*}[t]
    \centering
    \begin{tabular}{ c | c | c | c | c | c }
        \toprule
        Sequence 
        & ORB-SLAM3~\cite{camposORBSLAM3AccurateOpenSource2021} 
        & OKVIS2-X~\cite{bocheOKVIS2XOpenKeyframebased2025}
        & OKVIS2-X-Unimatch~\cite{bocheOKVIS2XOpenKeyframebased2025}
        & VIO-KLT (proposed)
        & VIO-SL (proposed) \\
        \hline
        Conveyor1
        & \textcolor{gray}{99.3} / 0.825 / 2.404
        & \textcolor{gray}{100.} / 0.734 / 2.068
        & \textcolor{gray}{100.} / \textcolor{red}{0.414} / \textcolor{blue}{1.100}
        & \textcolor{gray}{100.} / 0.455 / 1.605
        & \textcolor{gray}{100.} / \textcolor{red}{\textbf{0.332}} / \textcolor{blue}{\textbf{1.076}} \\
        Conveyor2
        & \textcolor{gray}{96.5} / 1.132 / 5.320
        & \textcolor{gray}{100.} / 1.944 / 6.753
        & \textcolor{gray}{100.} / \textcolor{red}{0.981} / \textcolor{blue}{4.082}
        & \textcolor{gray}{8.99} / - / -
        & \textcolor{gray}{100.} / \textcolor{red}{\textbf{0.347}} / \textcolor{blue}{\textbf{1.437}} \\
        Truck1
        & \textcolor{gray}{94.6} / \textcolor{red}{\textbf{0.215}} / \textcolor{blue}{\textbf{0.682}}
        & \textcolor{gray}{100.} / 0.533 / 1.767
        & \textcolor{gray}{100.} / 0.752 / \textcolor{blue}{1.380}
        & \textcolor{gray}{100.} / \textcolor{red}{0.370} / 1.442
        & \textcolor{gray}{100.} / 0.530 / 1.870 \\
        Truck2
        & \textcolor{gray}{90.6} / 0.384 / 2.002
        & \textcolor{gray}{99.8} / 0.724 / 1.577
        & \textcolor{gray}{99.8} / 0.966 / 1.643
        & \textcolor{gray}{99.9} / \textcolor{red}{\textbf{0.211}} / \textcolor{blue}{\textbf{0.889}}
        & \textcolor{gray}{99.9} / \textcolor{red}{{0.233}} / \textcolor{blue}{0.909} \\
        \bottomrule
    \end{tabular}
    \caption{
        The ground truth coverage and position and rotation ATE are shown in the format of ${(\SI{}{\percent} / \SI{}{\meter} / \SI{}{\degree})}$ for all sequences.
        The ATEs are computed over the overlapping regions.
        The lowest translation and rotation errors are highlighted in bolded \textcolor{red}{\textbf{red}} and \textcolor{blue}{\textbf{blue}}, and the runner-up in just \textcolor{red}{red} and \textcolor{blue}{blue}.
        Failed algorithms with less than 70\% ground truth coverage are omitted from the table.
        The proposed system outperforms the existing solutions.
    }
    \label{tab:vio_ate}
\end{table*}
\begin{figure*}[t]
    \centering
    \includegraphics[width=\linewidth]{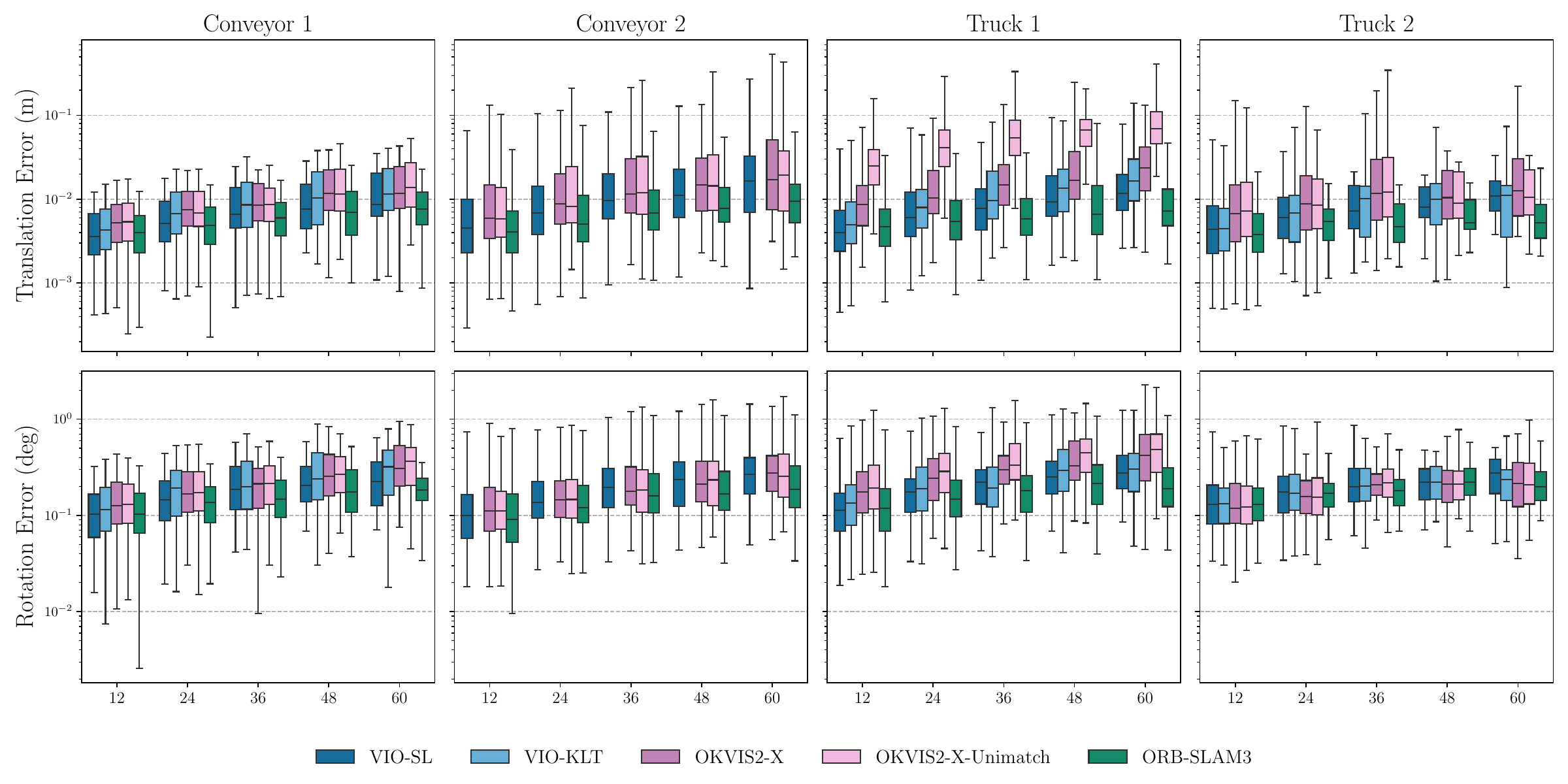}
    \caption{
        Relative trajectory error results of visual-inertial algorithms on the Quarry dataset shown in a log scale.
        The position and rotation RTE for all sequences are shown.
        The number of frames are shown in the x-axis as the segment lengths.
        Failed algorithms with less than 70\% ground truth coverage are omitted.
        The proposed method outperforms OKVIS2-X and performs on par with ORB-SLAM3.
    }
    \label{fig:vio_rte}
\end{figure*}

\subsubsection{Trajectory Alignment and Evaluation}
\label{subsubsec:trajectory_eval_alignment}
As described in Section~\ref{subsec:comp-algos}, the base configuration of the state-of-the-art algorithms is modified until the accuracy is saturated to the best of the authors' ability. Each algorithm is run multiple times, and the best performing run is reported.
Each algorithm may initialize and abort at any point in time of the experiment resulting in different lengths of estimated states.
Thus, the ground truth coverage is computed by the total duration of the estimated trajectory over the duration of the ground truth trajectory.
This metric represents the algorithm's robustness to run in tough underwater scenarios.
The algorithms with less than ${70\%}$ of the ground truth coverage are considered as failure and are omitted from evaluation.
The absolute and relative trajectory errors are then computed using the common segments of the estimates.

The absolute errors are computed first by querying the ground truth states at the estimated timestamps and aligning the estimated trajectories to the ground truth using the largest common trajectory segment~\cite{Umeyama1991}.
Then, the absolute trajectory error is computed by
\begin{align}
    e_{\textrm{ATE}} = \sqrt{\f{1}{K}\sum_{k=1}^{K} \Vert \bar{\mbf{T}}_{ab_k} \ominus \hat{\mbf{T}}_{ab_k} \Vert_{2}^{2}},
\end{align}
%
where ${K}$ is the number of pose measurements, ${\bar{(\cdot)}}$ denotes an aligned ground truth quantity, and ${\hat{(\cdot)}}$ denotes an aligned estimated quantity.

Relative trajectory errors are computed by first finding the common trajectory segment over the succeeded algorithms and computing the relative error over multiple segments using evo~\cite{grupp2017evo}.
Relative errors with smaller segment lengths represent local consistency, while the errors with larger segment lengths reflect the long-term accuracy~\cite{Zhang2018Tutorial}. For $N_s$ segments from a chosen segment length $n_s$, the relative error at each segment $k \in [1, N_s]$ is given by
\begin{align}
    e_{\textrm{RTE},k} = \Vert \bar{\mbf{T}}^{\Delta} \ominus \hat{\mbf{T}}^{\Delta} \Vert_2,
\end{align}
where ${\mbf{T}^{\Delta} = \mbf{T}_{ab_{k+n_s}} \ominus \mbf{T}_{ab_k}}$.
The non-overlapping segment lengths are chosen from 2\% of the total duration to 10\% in an increment of 2\% of the longest sequence, Conveyor2 (\SI{26}{\minute}).
Time is used as the segmentation metric here to measure an algorithm's ability to track a vehicle at rest and in motion. Unlike an indoor setting, stationary motion is no guarantee of accurate tracking in subsea environments, as visual saliency varies with time even for the same scene. For example, visual-inertial algorithms are likely to drift more when the ROV is far from a target with poor visibility even at rest, since the localization is heavily relying on an IMU.
\vspace{0.1cm}

\subsubsection{Visual-inertial Evaluation}
\label{subsubsec:results-experimental-visual-inertial}
The output trajectories of the visual-inertial methods are shown in Fig.~\ref{fig:vio_traj}.
The estimated trajectories of a better performing method between the VIO-SL/VIO-KLT, a better performing method between OKVIS2-X/OKVIS2-X-Unimatch, and ORB-SLAM3 are plotted along with the ground truth trajectories from the photogrammetry model shown in black.

The proposed VIO-SL (blue) is the only method to detect repeatable features at the beginning of Conveyor1, where the ROV undergoes erratic motion at a high distance from a visual asset. As such, it is the only method able to provide reasonable estimates for that period. For Conveyor2, which is the most challenging out of all four due to its high motion profile and repetitive texture, only the results of the proposed method align with the ground truth. Note that the proposed architecture does not contain any place recognition or loop closure module to form long-term data associations, and the competing algorithms that do, OKVIS2-X and ORB-SLAM3, are ineffective at correcting their long-term drift on this sequence despite finding loop closures. Specifically, OKVIS2-X rejects many loop closures due to indistinctive feature descriptors.

ORB-SLAM3 outperforms the proposed system on Truck1, but is unable to estimate around 80 seconds of the whole trajectory, during particularly difficult regions with a high distance to visual asset. This is largely due to the difficulty of matching ORB features at high distances in an unstructured setting. OKVIS2-X-Unimatch shows poorer performance in the longer-term RTE due to much of Truck1 being flat seabed and not geometrically distinctive. The Truck2 sequence has a high amount of visual floaters and low distance to visual asset, and the proposed method is able to outperform the competing methods due to the robustness in the learned matching process. ORB-SLAM3 suffers consistent visual aliasing in its loop closure pipeline as the asset contains many visually similar structures, and has notably low trajectory coverage due to the initial period being almost all water column and unstructured seabed. Additionally, the tracking pipelines in both ORB-SLAM3 and OKVIS2-X are consistently distracted by sand clouds thrown up by the ROV, causing erroneous tracking of dynamic particles.
\begin{figure*}[t]
    \centering
    \begin{subfigure}[t]{0.5\linewidth}
        \centering
        \includegraphics[width=\linewidth]{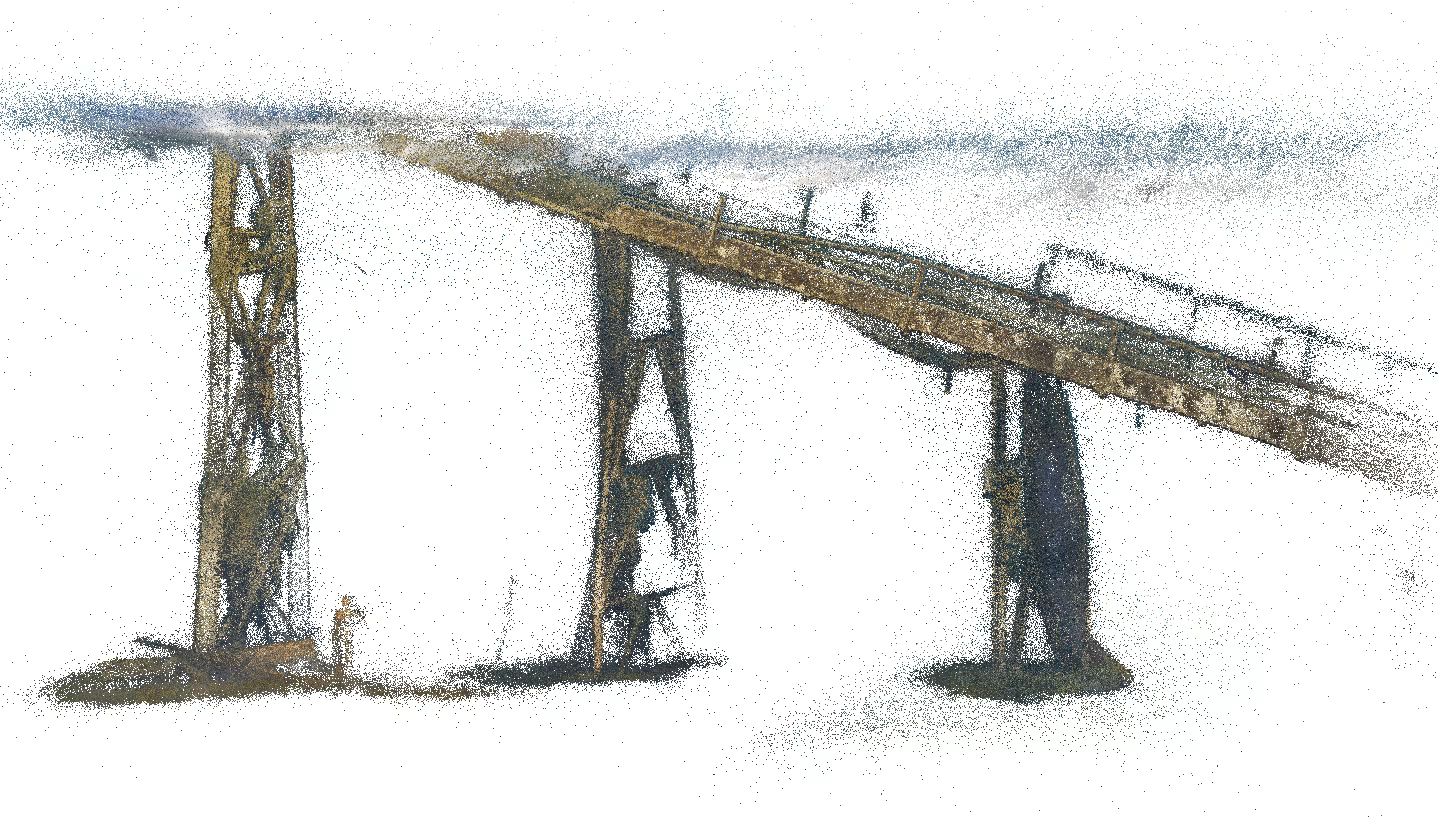}
        \caption{Agisoft Metashape.}
        \label{fig:sparse_conv_1_gt}
    \end{subfigure}%
    \begin{subfigure}[t]{0.5\linewidth}
        \centering
        \includegraphics[width=\linewidth]{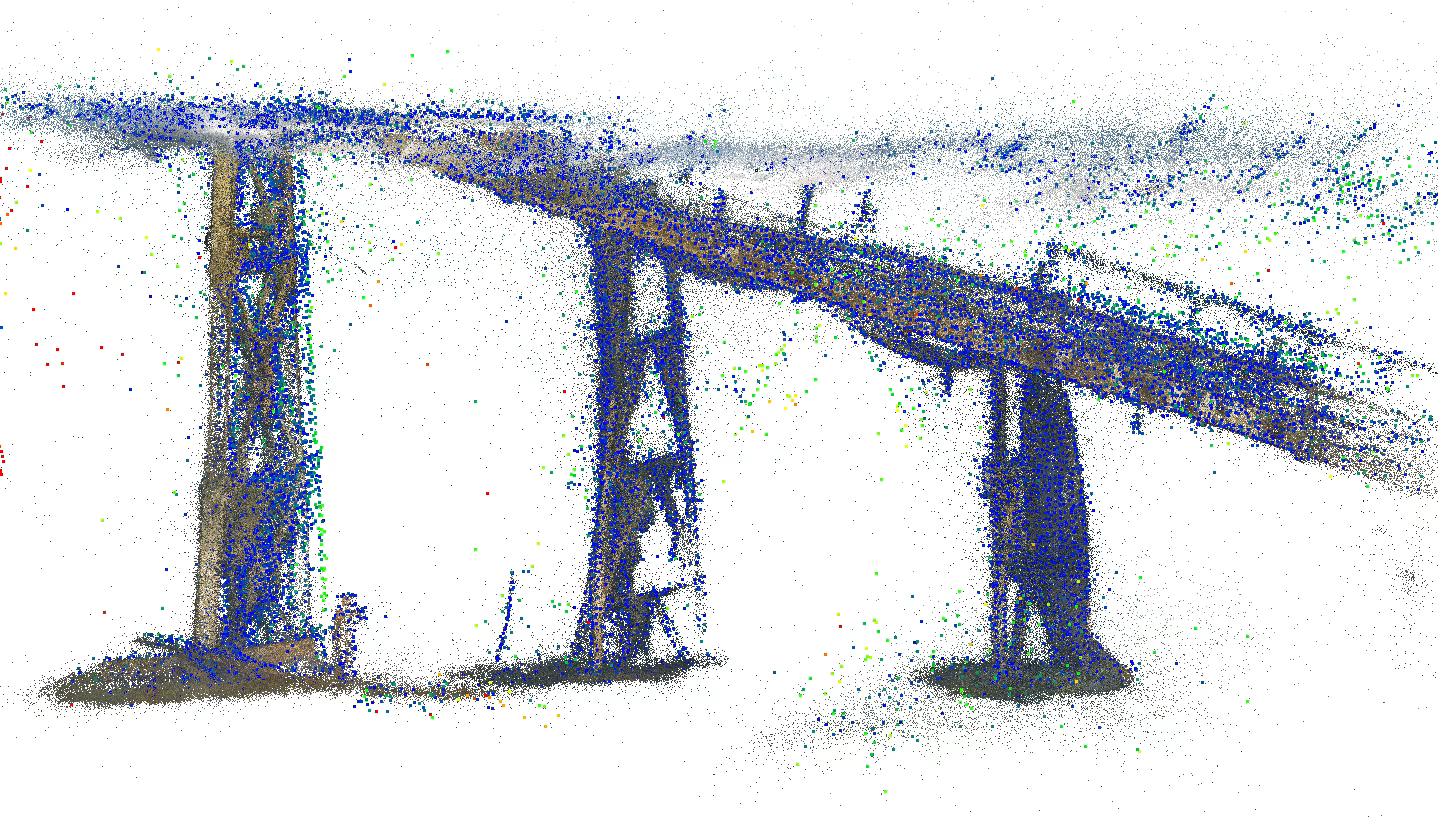}
        \caption{VIO-SL.}
        \label{fig:sparse_conv_1_ours}
    \end{subfigure}
    \vskip\baselineskip
    \begin{subfigure}[t]{0.5\linewidth}
        \centering
        \includegraphics[width=\linewidth]{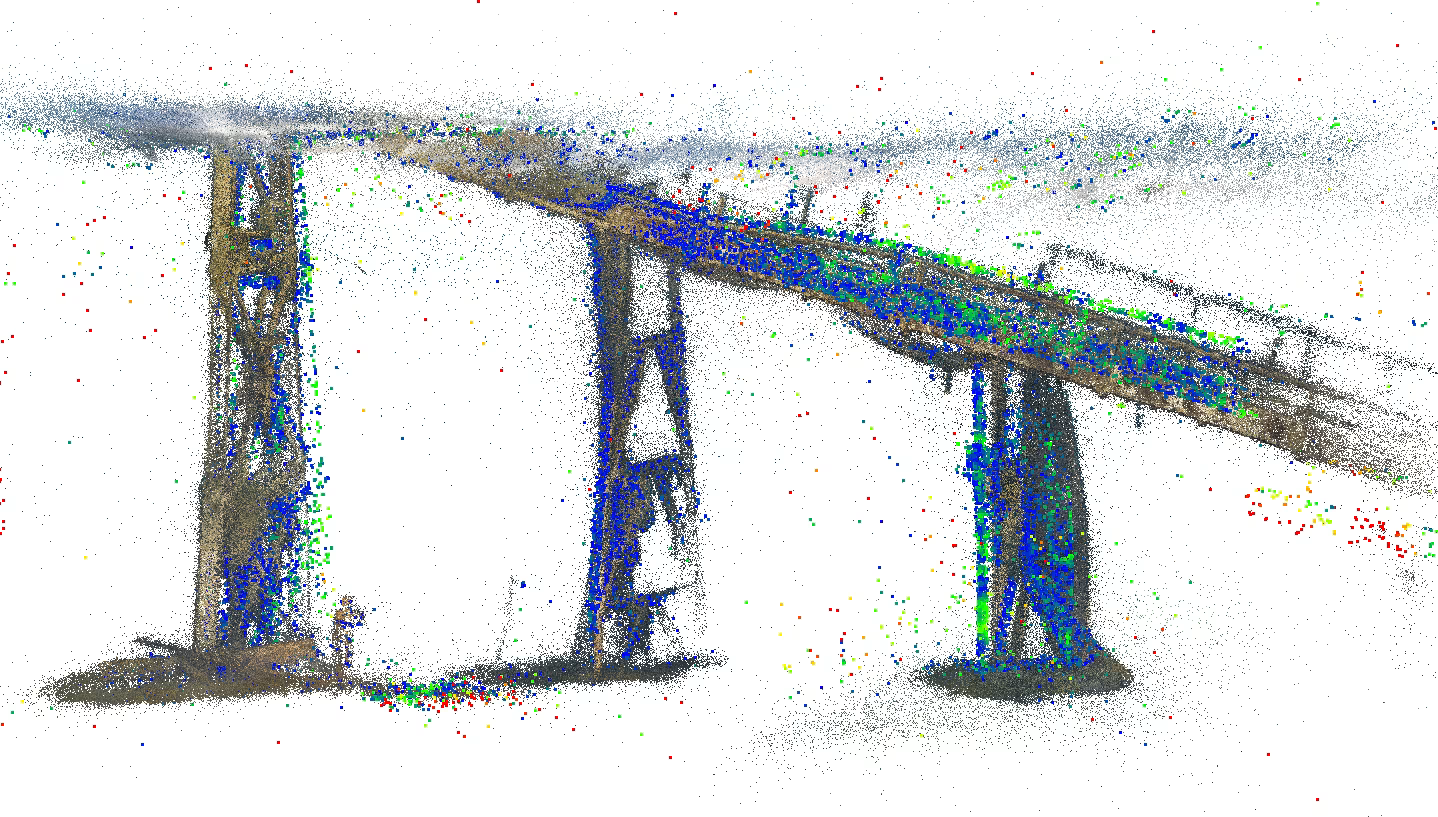}
        \caption{ORB-SLAM3~\cite{camposORBSLAM3AccurateOpenSource2021}.}
        \label{fig:sparse_conv_1_orbslam3}
    \end{subfigure}%
    \begin{subfigure}[t]{0.5\linewidth}
        \centering
        \includegraphics[width=\linewidth]{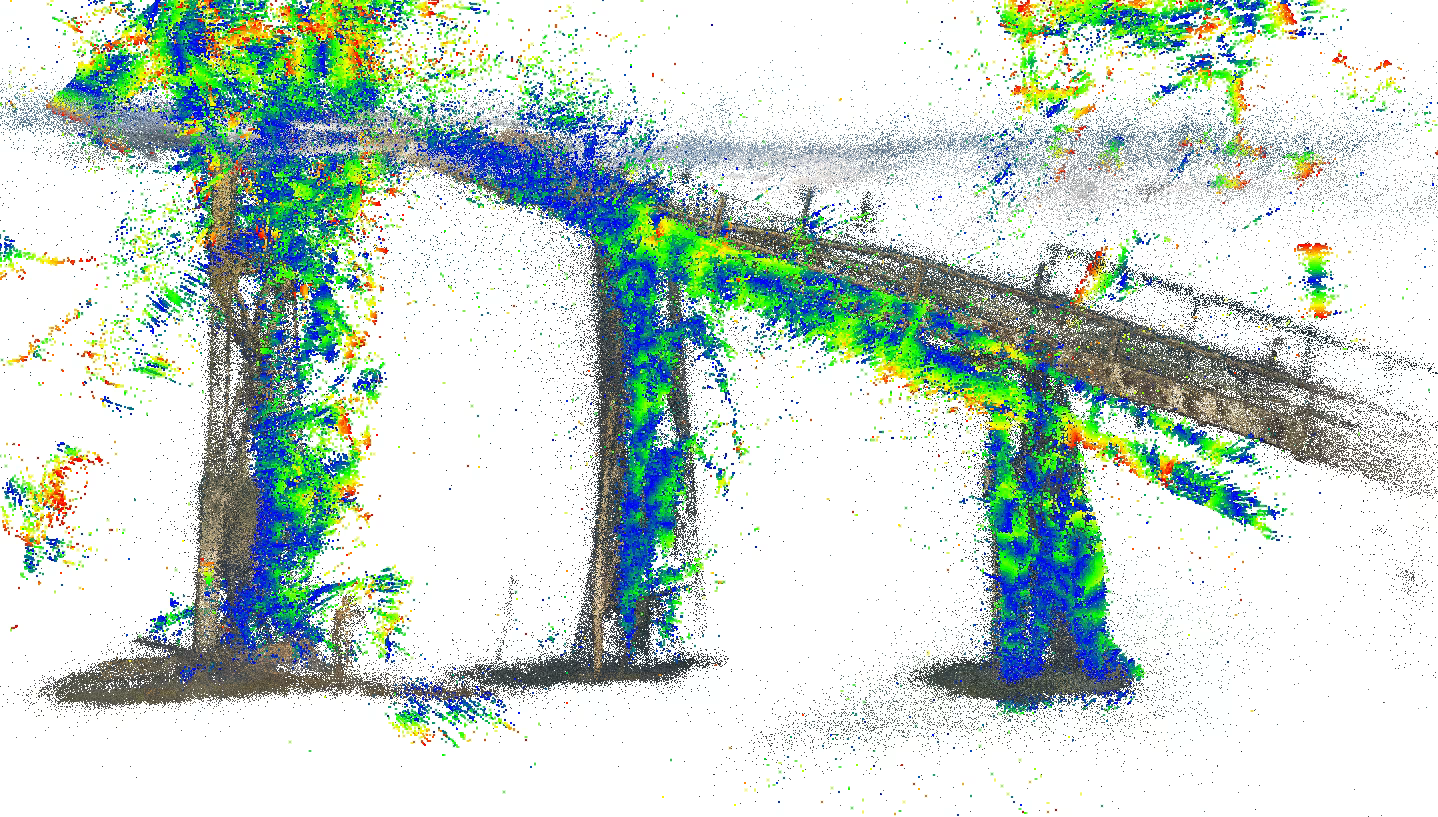}
        \caption{OKVIS2-X~\cite{bocheOKVIS2XOpenKeyframebased2025} with Unimatch~\cite{xuUnifyingFlowStereo2023} submapping.}
        \label{fig:sparse_conv_1_okvis2x}
    \end{subfigure}
    \caption{
        Photogrammetric Conveyor1 sparse feature cloud. The point distances between the evaluated visual-inertial methods and the ground truth are shown with the same color scale throughout.
        While the state-of-the-art methods show either poor coverage, significant delamination, or skew, the proposed method is able to show significant coverage without sacrificing accuracy.
    }
    \label{fig:vio_sparsemap}
\end{figure*}
The ground truth coverage and absolute trajectory errors of the visual-inertial algorithms are shown in Table~\ref{tab:vio_ate} in the format of ${(\SI{}{\percent} / \SI{}{\meter} / \SI{}{\degree})}$ with the lowest translation and rotation errors highlighted in bolded \textcolor{red}{\textbf{red}} and \textcolor{blue}{\textbf{blue}}, and the runner-up unbolded \textcolor{red}{red} and \textcolor{blue}{blue}.
The initial proximity to an asset in Conveyor2 is far, resulting in 3.5\%, or \SI{55}{\second}, coverage loss in ORB-SLAM3 and tracking failure in the VIO-KLT.
However, the proposed SL-extension still outperforms in both position and rotation ATE, precisely, by 20\% and 65\%, respectively, over their corresponding runner-ups. 
ORB-SLAM3 has the lowest overall coverage due to consistent failure under visually challenging conditions and a conservative re-initialization strategy.
For instance, the 9.4\%, or $\SI{54}{\second}$, coverage loss for ORB-SLAM3 is notable for Truck2 due to the generally high proximity to target.
The initial untracked period is shown in Fig.~\ref{fig:vio_traj_truck2_init}, where only the proposed system is able to provide accurate state estimates for the low proximity trajectory segment, which is approximately 56 seconds long.

The relative trajectory errors for the visual-inertial methods are shown in Fig.~\ref{fig:vio_rte}.
The proposed VIO-SL performs considerably better in both translation and rotation relative errors for all segment lengths against the KLT and OKVIS2-X and comparably to ORB-SLAM3 in smaller trajectory lengths.
While ORB-SLAM3 outperforms at higher trajectory lengths, it should be noted that as it had the lowest coverage, all algorithms were truncated to the ORB-SLAM3 trajectory segment.
Additionally, it showed high individual run variance, failing on Truck2 for 60\% of total runs.
As such, despite ORB-SLAM3 being accurate in visually salient regions with low proximity and structured environments, the proposed system is more robust and performs better in visually challenging scenarios, as the learned extractor is able to localize unstructured features even at high distances.
As a corollary, the sparse coverage cloud is shown in Fig.~\ref{fig:vio_sparsemap}, where despite marginally outperforming the proposed system in long-term RTE, ORB-SLAM3 shows low overall asset coverage.
\begin{figure*}[t]
    \centering
    \includegraphics[width=0.5\linewidth]{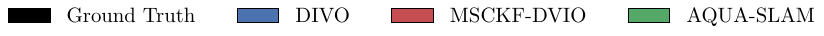}
    \begin{subfigure}[t]{0.5\linewidth}
        \centering
        \includegraphics[width=\linewidth]{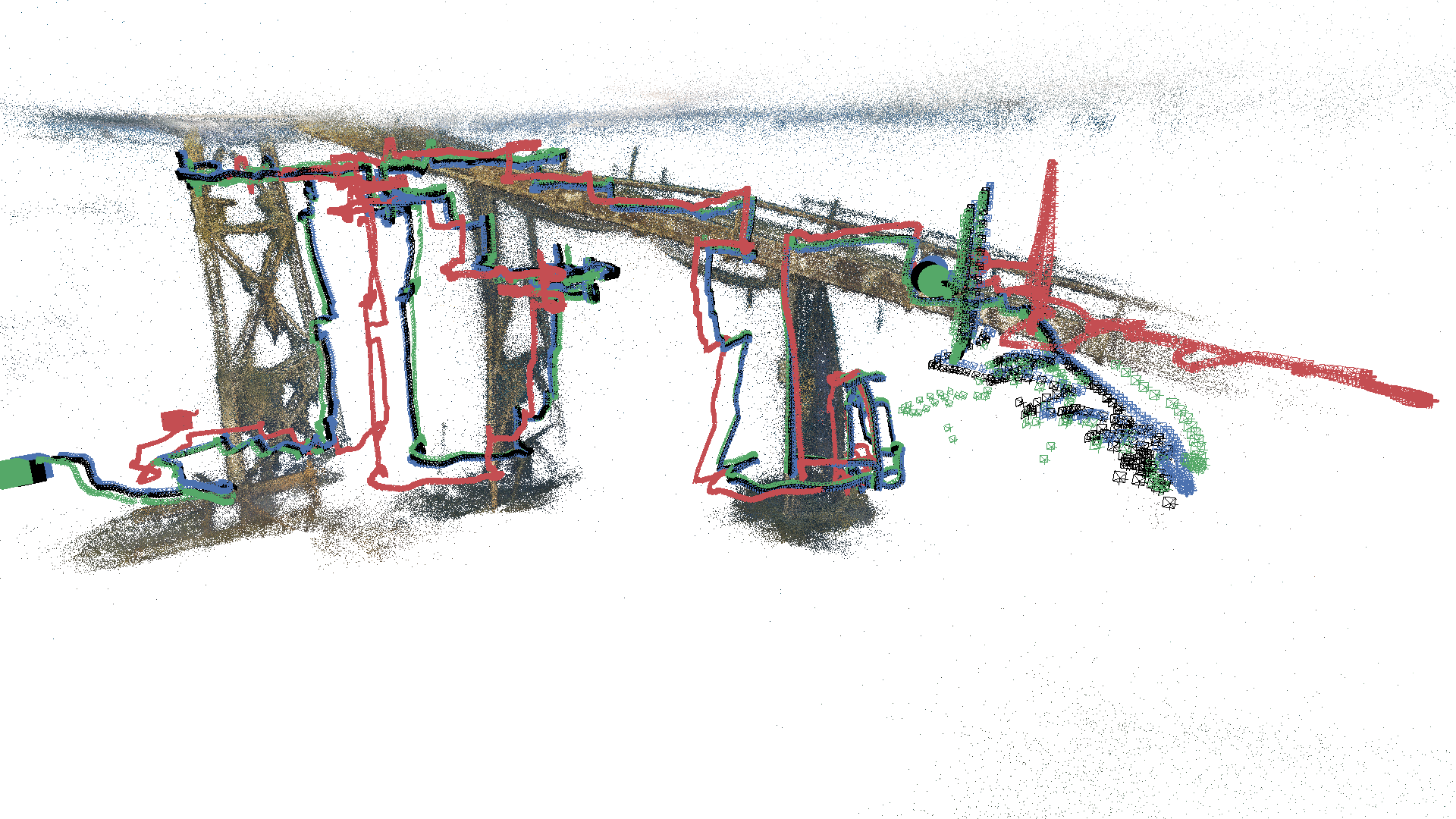}
        \caption{Conveyor1.}
        \label{fig:divo_traj_conv1}
    \end{subfigure}%
    \begin{subfigure}[t]{0.5\linewidth}
        \centering
        \includegraphics[width=\linewidth]{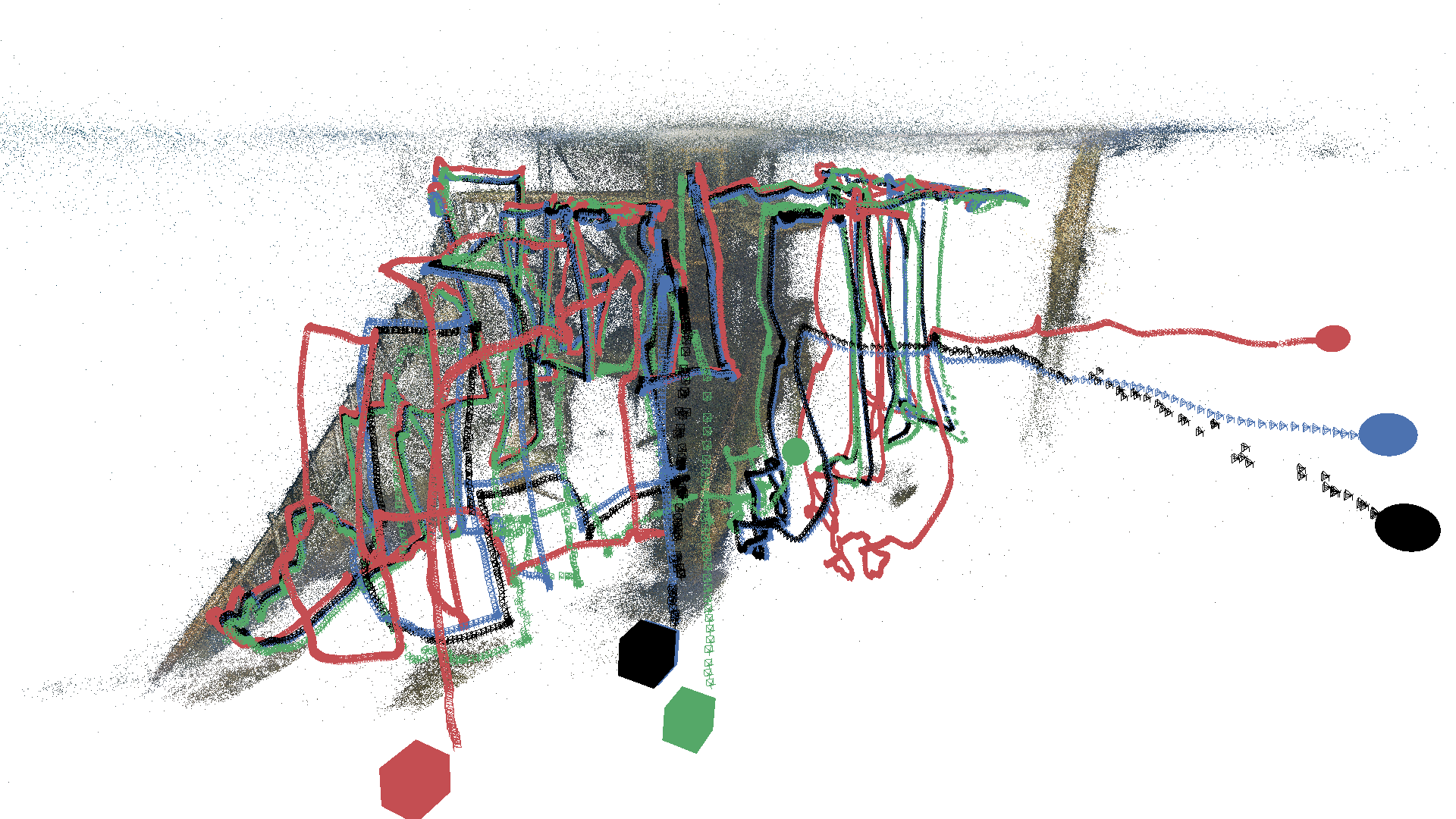}
        \caption{Conveyor2.}
        \label{fig:divo_traj_conv2}
    \end{subfigure}
    \vskip\baselineskip
    \begin{subfigure}[t]{0.5\linewidth}
        \centering
        \includegraphics[width=\linewidth]{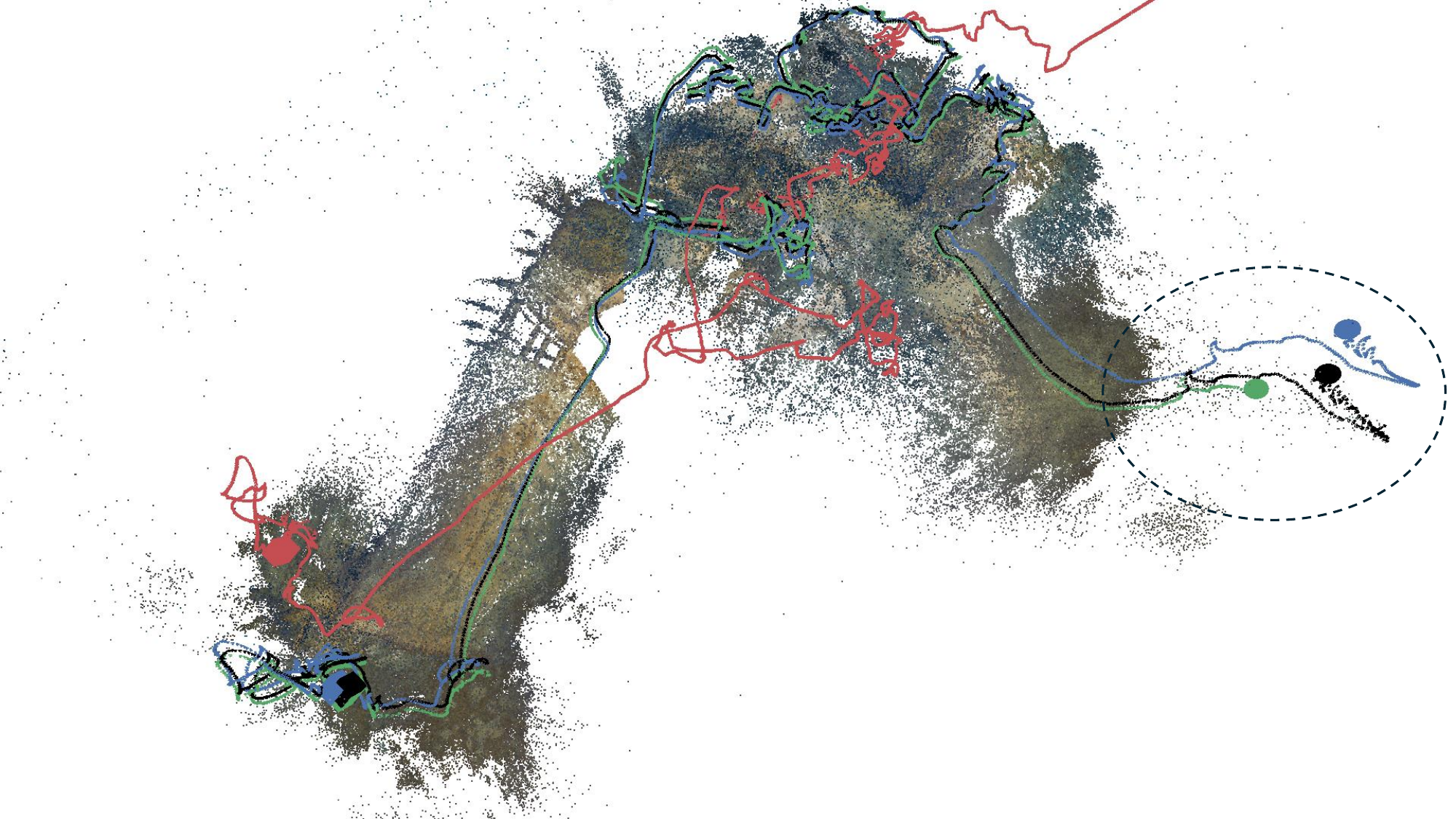}
        \caption{Truck1.}
        \label{fig:divo_traj_truck1}
    \end{subfigure}%
    \begin{subfigure}[t]{0.5\linewidth}
        \centering
        \includegraphics[width=\linewidth]{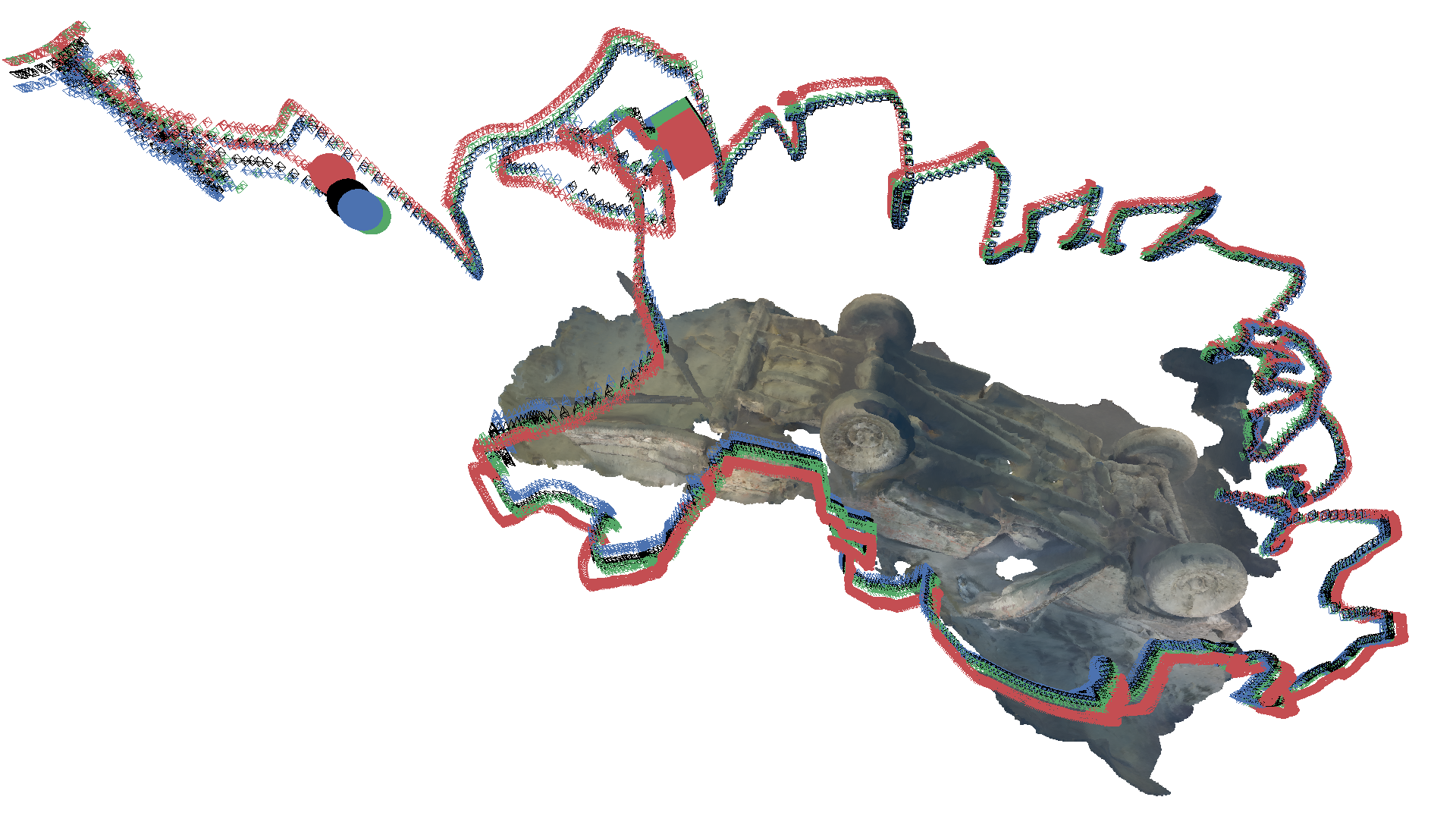}
        \caption{Truck2.}
        \label{fig:divo_traj_truck2}
    \end{subfigure}
    \vskip\baselineskip
    \begin{subfigure}[t]{\linewidth}
        \centering
        \includegraphics[width=.4\linewidth]{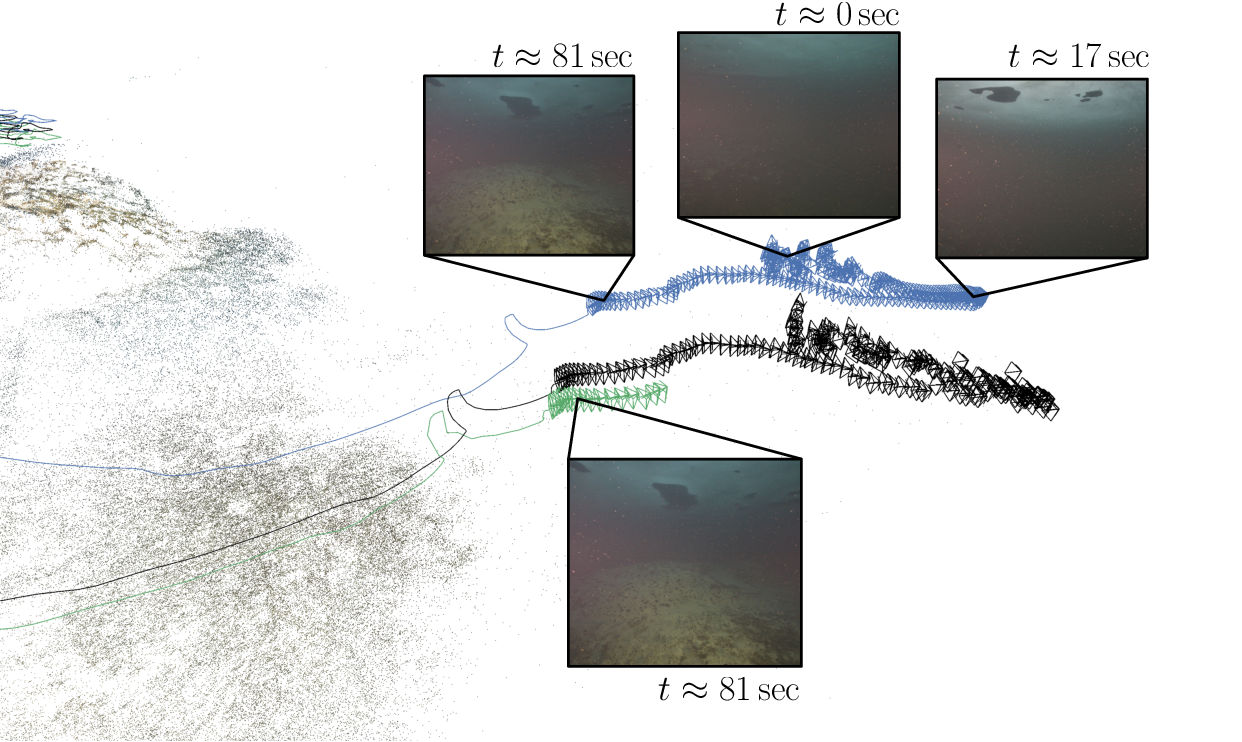}
        \caption{DIVO (\textcolor{blue}{blue}) vs. AQUA-SLAM (\textcolor{teal}{green}) in the initial region of Truck1, circled in (c). AQUA-SLAM is unable to track for the first $\approx \SI{81}{\sec}$.}
        \label{fig:divo_traj_truck1_init}
    \end{subfigure}
    \caption{
        The trajectory of acoustic-visual-inertial algorithms overlaid to the ground truth generated by Agisoft Metashape (black).
        The monocular-based MSCKF-DVIO performs poorly in all sequences.
        The proposed DIVO (blue) shows highest agreement with the ground truth. Spheres denote starting positions and boxes denote ending positions for each method.
    }
    \label{fig:divo_traj}
\end{figure*}
Overall, the proposed VIO-SL achieves complete trajectory and asset coverage despite low visibility and photometric disturbances. Although ORB-SLAM3 outperforms in long-term RTE, it shows conservative trajectory estimates and high stochasticity due to tracking failure and visual aliasing, making the proposed system more suitable for real-time deployment.
\vspace{0.1cm}

\subsubsection{Acoustic-visual-inertial Evaluation}
\label{subsubsec:results-experimental-acoustic-visual-inertial}
With the exceeding performance of the SuperPoint~\cite{detoneSuperPointSelfSupervisedInterest2018} and LightGlue~\cite{lindenbergerLightGlueLocalFeature2023} extension over the KLT method in both short-term and long-term accuracy, the DVL is only extended to the VIO with the learned frontend.
\renewcommand{\arraystretch}{1.2}
\begin{table*}[t]
    \centering
    \begin{tabular}{ c | c | c | c}
        \toprule
        Sequence 
        & MSCKF-DVIO~\cite{Zhao2023Tightly} 
        & AQUA-SLAM~\cite{Xu2025AQUA-SLAM} 
        & DIVO (proposed) \\
        \hline
        Conveyor1
        & \textcolor{gray}{99.6} / 0.912 / 3.834
        & \textcolor{gray}{99.6} / \textcolor{red}{0.344} / \textcolor{blue}{1.756}
        & \textcolor{gray}{100.} / \textcolor{red}{\textbf{0.258}} / \textcolor{blue}{\textbf{0.964}} \\
        Conveyor2
        & \textcolor{gray}{100.} / 0.757 / 3.764
        & \textcolor{gray}{97.1} / \textcolor{red}{0.534} / \textcolor{blue}{2.536}
        & \textcolor{gray}{100.} / \textcolor{red}{\textbf{0.271}} / \textcolor{blue}{\textbf{1.050}} \\
        Truck1
        & \textcolor{gray}{97.9} / 2.098 / 6.930
        & \textcolor{gray}{96.1} / \textcolor{red}{\textbf{0.451}} / \textcolor{blue}{\textbf{1.482}}
        & \textcolor{gray}{100.} / \textcolor{red}{0.507} / \textcolor{blue}{1.688} \\
        Truck2
        & \textcolor{gray}{100.} / 0.389 / 1.199
        & \textcolor{gray}{99.9} / \textcolor{red}{0.258} / \textcolor{blue}{1.135}
        & \textcolor{gray}{99.9} / \textcolor{red}{\textbf{0.229}} / \textcolor{blue}{\textbf{0.972}} \\
        \bottomrule
    \end{tabular}
    \caption{
        The ground truth coverage and position and rotation ATE are shown in the format of ${(\SI{}{\percent} / \SI{}{\meter} / \SI{}{\degree})}$ for all sequences.
        The ATEs are computed over the overlapping regions.
        The best position and rotation ATEs are bolded and highlighted, and the runner ups are only highlighted in red and blue, respectively, for each sequence.
        The proposed system outperforms the existing solutions.
    }
    \label{tab:divo_ate}
\end{table*}
\begin{figure*}[h!]
    \centering
    \includegraphics[width=\linewidth]{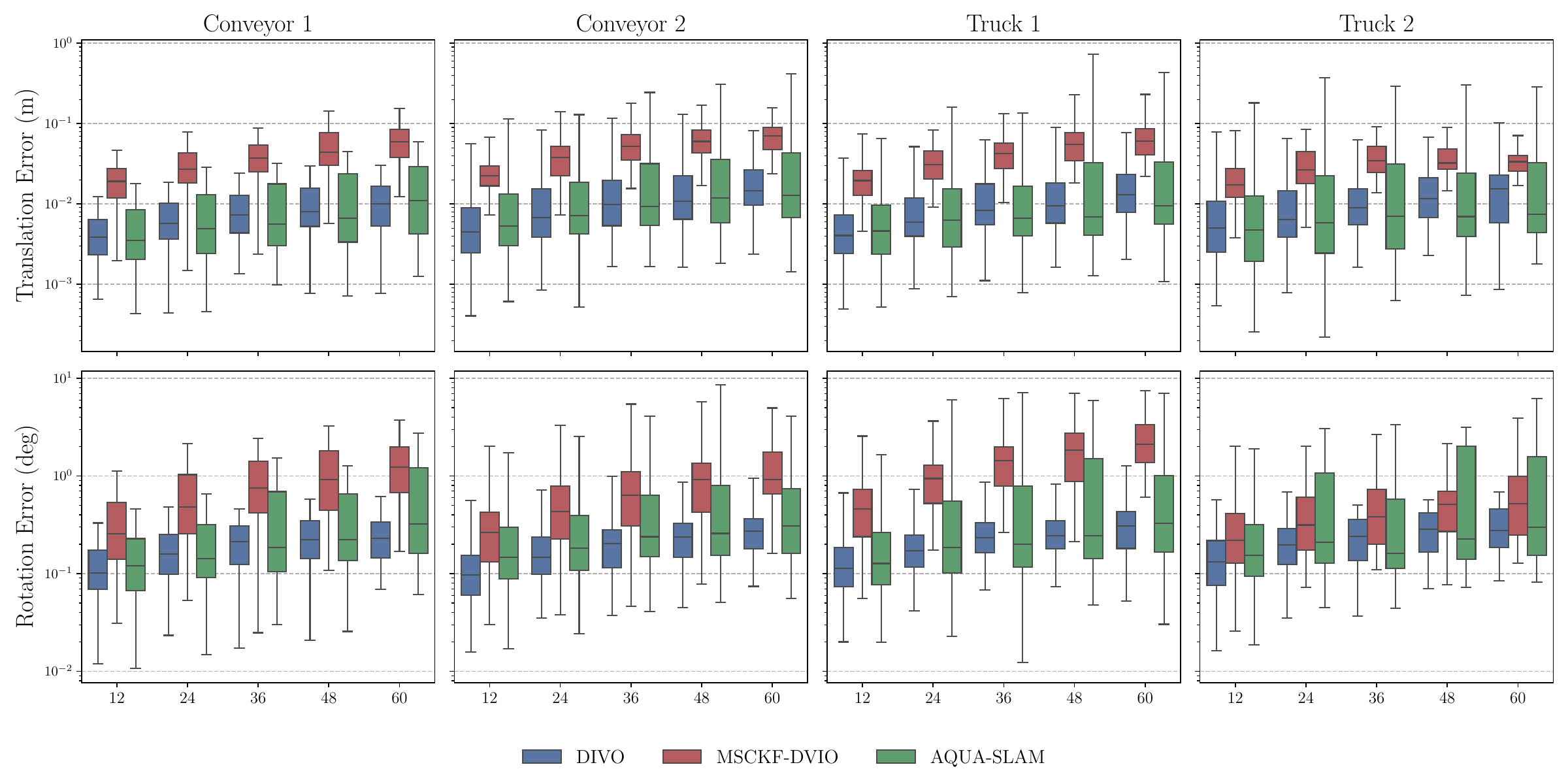}
    \caption{
        Relative trajectory error results of acoustic-visual-inertial algorithms on the Quarry dataset.
        The position and rotation RTE for all sequences are shown.
        The number of frames are shown in the x-axis as the segment lengths.
        The proposed method with the SL outperforms both MSCKF-DVIO and AQUA-SLAM.
    }
    \label{fig:divo_rte}
\end{figure*}

The output trajectories of the acoustic-visual-inertial methods are shown in Fig.~\ref{fig:divo_traj}.
The estimated trajectories from the competing algorithms are overlaid on the ground truth for qualitative comparison.
Overall, the proposed method and AQUA-SLAM estimate the trajectory more accurately than MSCKF-DVIO.
The drawbacks of using a monocular camera cannot be fully compensated for by incorporating a DVL, and it does not perform on par with the stereo systems.
For instance, the Umeyama method~\cite{Umeyama1991} is not able to align the trajectory from MSCKF-DVIO for Truck1 due to the erroneous estimates.
Although the filtering-based monocular-DVL method is able to provide qualitatively tractable results for Truck2, the optimization-based approaches, namely the proposed system and AQUA-SLAM, significantly outperform.

Existing DVL-aided methods claim their robustness to visual tracking failure by employing the acoustic sensor, thus, the ground truth coverage is expected to be higher than those of visual-inertial methods.
As claimed, AQUA-SLAM, which is based on ORB-SLAM3, has a higher ground truth coverage over all sequences than its precedent.
The difference is most pronounced in Truck2.
However, AQUA-SLAM still does not achieve full coverage of the ground truth for all sequences, which may be due to the inherited conservative initialization structure of ORB-SLAM3. An example of this is shown in Fig.~\ref{fig:divo_traj_truck1_init}, where AQUA-SLAM does not produce a trajectory estimate for the first 81 seconds of Truck1 until closer proximity to asset is reached.
Also, at the beginning of the Conveyor1 where the ROV is at a high distance to the target, AQUA-SLAM camera estimates are unstable and unaligned with the ground truth, while the proposed method (blue) aligns with the ground truth.

The absolute trajectory results are presented in Table~\ref{tab:divo_ate} in the same format as the visual-inertial case.
The lowest translation and rotation errors are highlighted in bolded \textcolor{red}{\textbf{red}} and \textcolor{blue}{\textbf{blue}}, and the runner-up unbolded \textcolor{red}{red} and \textcolor{blue}{blue}.
The ATE results of the MSCKF-DVIO are considerably higher than the optimization-based approaches and confirms the qualitative analysis.
The proposed method achieves 25\% and 45\% decrease in position and rotation ATE, respectively, for Conveyor1, and 49\% and 59\% for Conveyor2.
Incidentally, DIVO, the proposed DVL-aided solution, outperforms its visual-inertial equivalent in visually challenging Conveyor1 and Conveyor2 by (22\% / 10\%) and (22\% / 27\%) in position and rotation ATEs, respectively.
AQUA-SLAM marginally outperforms the proposed system on Truck1, however, the inclusion of DVL to the system closes the gap from \SI{31.5}{\centi\meter} in the visual-inertial case to \SI{5.6}{\centi\meter}.

\begin{figure*}[t]
    \centering
    \begin{subfigure}[t]{0.33\linewidth}
        \centering
        \includegraphics[width=\linewidth]{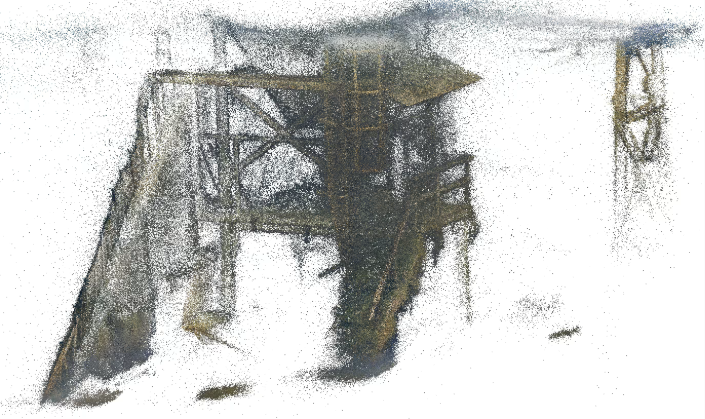}
        \caption{Agisoft Metashape.}
        \label{fig:sparse_conv_2_gt}
    \end{subfigure}%
    \begin{subfigure}[t]{0.33\linewidth}
        \centering
        \includegraphics[width=\linewidth]{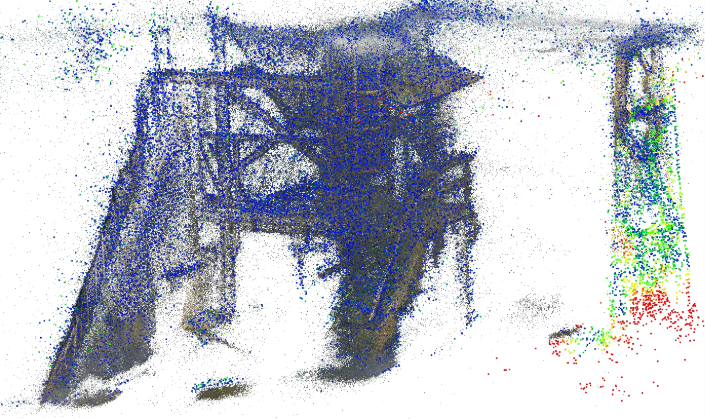}
        \caption{DIVO.}
        \label{fig:sparse_conv_1_ours}
    \end{subfigure}%
    \begin{subfigure}[t]{0.33\linewidth}
        \centering
        \includegraphics[width=\linewidth]{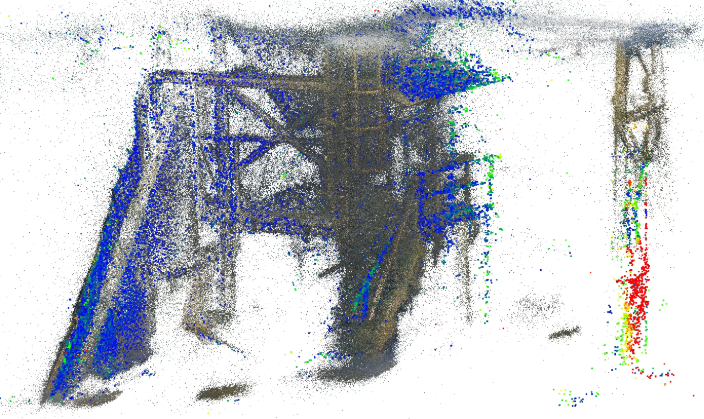}
        \caption{AQUA-SLAM~\cite{Xu2025AQUA-SLAM}.}
        \label{fig:sparse_conv_2_aquaslam}
    \end{subfigure}%
    \caption{
        The point distances between the DIVO and AQUA-SLAM~\cite{Xu2025AQUA-SLAM} sparse feature cloud and the ground truth are shown here. The proposed method has significantly higher asset coverage.
    }
    \label{fig:divo_sparsemap}
\end{figure*}

The relative trajectory errors for the acoustic-visual-inertial methods are shown in Fig.~\ref{fig:divo_rte}.
The proposed DIVO performs considerably better in both translation and rotation relative errors for all segment lengths in all sequences.
Although AQUA-SLAM has a lower median translation for longer segment lengths in Truck1 and Truck2, the median difference between the two visual-inertial equivalents is narrowed down.
More importantly, lower variance in the proposed method signifies that it is more robust to challenging scenes.
This is further proven in Fig.~\ref{fig:divo_sparsemap}, where the proposed system has significantly higher asset coverage on the Conveyor2 sequence, indicating that it is able to continue fusing acoustic-visual data in segments where AQUA-SLAM is entirely dependent on the DVL.

Overall, the proposed DIVO and VIO-SL excel across all metrics, including ground truth coverage, accuracy, and robustness, demonstrating a considerable performance improvement compared to their best-performing counterparts, ORB-SLAM3 and AQUA-SLAM. It should be noted that the proposed system is causal, only forming visual associations in a sliding window of $\SI{3}{\second}$. Despite this, the learned frontend allows the proposed system to find repeatable feature associations up to $\SI{10}{\meter}$ away from the vehicle, and is robust to dynamic particles and illumination variance. This allows the proposed system to outperform its non-causal competitors, whose long and mid-term data associations are less effective in the subsea domain due to repetitive structures and generally indistinctive features. Furthermore, the continuous-time graph optimization of DIVO allows other sensors to be seamlessly integrated as additional modules in the future.

\section{Conclusion}
\label{sec:conclusion}
This paper presents a novel continuous-time DVL-inertial-visual odometry (DIVO) system that integrates SuperPoint~\cite{detoneSuperPointSelfSupervisedInterest2018} features matched with LightGlue~\cite{lindenbergerLightGlueLocalFeature2023}.
The proposed system has been extensively tested both in simulation and real-world settings.
In simulation, a Monte Carlo analysis is performed to evaluate the accuracy, consistency, and computational efficiency of the proposed continuous-time backend framework.
The new framework is shown to yield higher accuracy at a cost of marginal increase in computation.
The consistency of the proposed backend system is validated in simulation by evaluating the average normalized estimation error squared (NEES).

DIVO is further tested in an experimental setting, where an ROV with pre-calibrated sensors is remotely operated for asset inspections. 
The proposed method is compared against a collection of state-of-the-art algorithms, which are tuned to the best of the authors' ability until the performance is saturated, to ensure a fair comparison.
All candidates are compared against the ground truth generated by photogrammetry and evaluated in terms of ground truth coverage, absolute trajectory error, and relative trajectory error.

For the first contribution of this paper, the proposed visual-inertial solution, VIO-SL, shows superior performance in visually challenging scenarios, achieving an accurate trajectory estimation with the highest trajectory and sparse map coverage.
Despite ORB-SLAM3 performing better in long-term relative error metrics, the proposed system is more robust and outperforms under visually challenging conditions.
The second contribution of this paper, DIVO, that builds upon the continuous-time backbone, excels in all of the evaluated metrics.
Furthermore, thanks to the nature of the continuous-time framework, the integration of other sensor modalities to the current pipeline is modular.
Our future work includes incorporating other sensor modalities and a loop closure module.









\printbibliography


\end{document}